\documentclass{bmvc2k}
\usepackage{comment}
\usepackage{amsmath}
\newcommand{\argmax}{\mathop{\rm arg~max}}

\title{Multi-scale Template Matching with Scalable Diversity Similarity in an Unconstrained Environment}

\addauthor{Yi Zhang}{y_zhang@scv.cis.iwate-u.ac.jp}{1}
\addauthor{Chao Zhang}{zhang@u-fukui.ac.jp}{2}
\addauthor{Takuya Akashi}{akashi@iwate-u.ac.jp}{3}

\addinstitution{
 Graduate School of Engineering\\
 Iwate University, Morioka, Japan
}
\addinstitution{
 Faculty of Engineering\\
 University of Fukui, Fukui city, Japan
}
\addinstitution{
 Faculty of Science and Engineering\\
 Iwate University, Morioka, Japan
}

\runninghead{Yi Zhang, Chao Zhang, Takuya Akashi}{Multi-scale Template Matching}

\begin{document}

\maketitle

\begin{abstract}
We propose a novel multi-scale template matching method which is robust against both scaling and rotation in unconstrained environments. The key component behind is a similarity measure referred to as scalable diversity similarity (SDS). Specifically, SDS exploits bidirectional diversity of the nearest neighbor (NN) matches between two sets of points. To address the scale-robustness of the similarity measure, local appearance and rank information are jointly used for the NN search. Furthermore, by introducing penalty term on the scale change, and polar radius term into the similarity measure, SDS is shown to be a well-performing similarity measure against overall size and rotation changes, as well as non-rigid geometric deformations, background clutter, and occlusions. The properties of SDS are statistically justified, and experiments on both synthetic and real-world data show that SDS can significantly outperform state-of-the-art methods.
\end{abstract}

\section{Introduction}
\label{sec:intro}
\begin{figure*}[t]
 \centering
 \subfigure[]{
  \label{fig:subfig:a} 
  \includegraphics[width=0.165\linewidth]{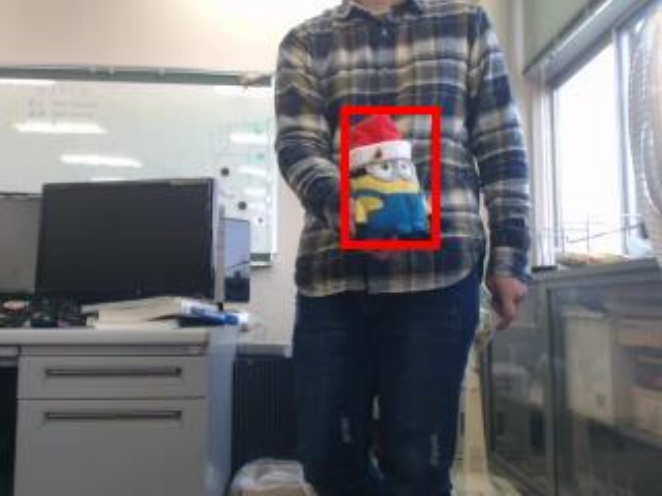}}
 \hspace{0.001in}
 \subfigure[]{
  \label{fig:subfig:b} 
  \includegraphics[width=0.165\linewidth]{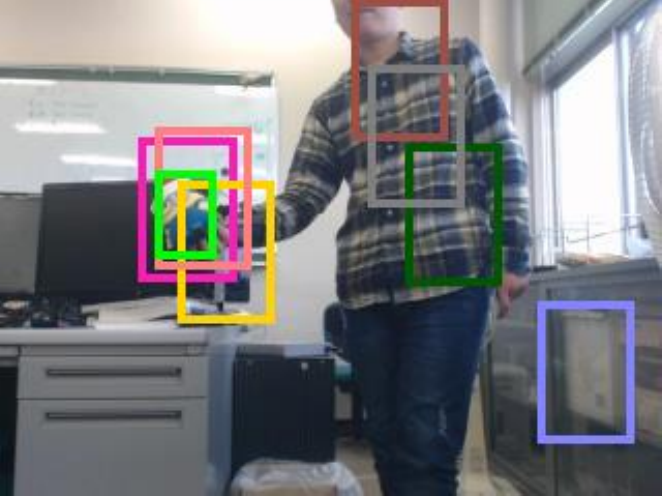}}
 \hspace{0.001in}
 \subfigure[]{
  \label{fig:subfig:c} 
  \includegraphics[width=0.165\linewidth]{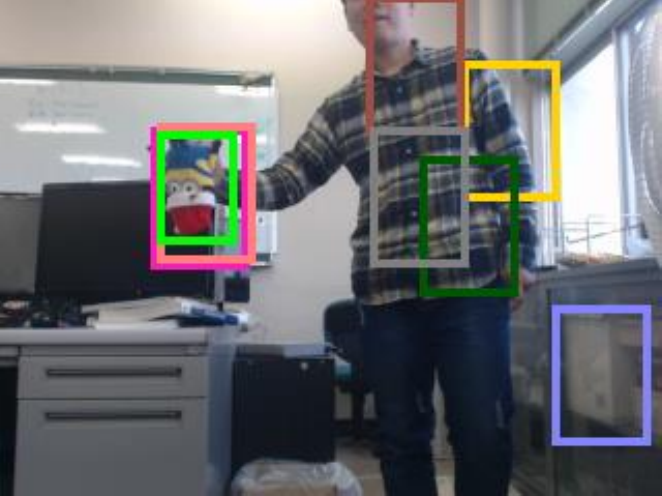}}
 \hspace{0.001in}
 \subfigure[]{
  \label{fig:subfig:d} 
  \includegraphics[width=0.165\linewidth]{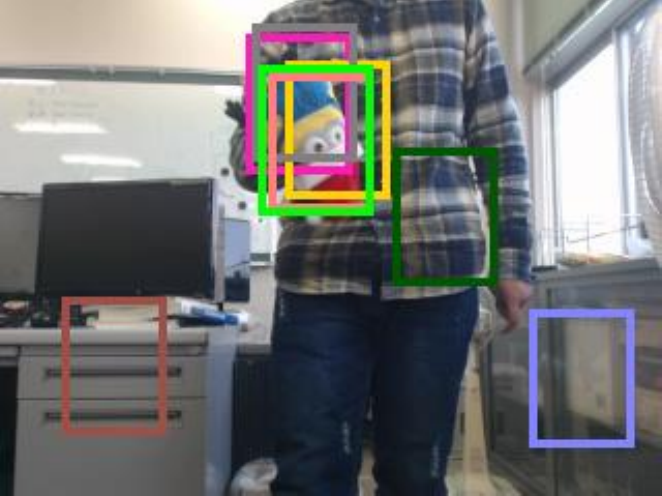}}
 \hspace{0.001in}
 \subfigure[]{
  \label{fig:subfig:e} 
  \includegraphics[width=0.165\linewidth]{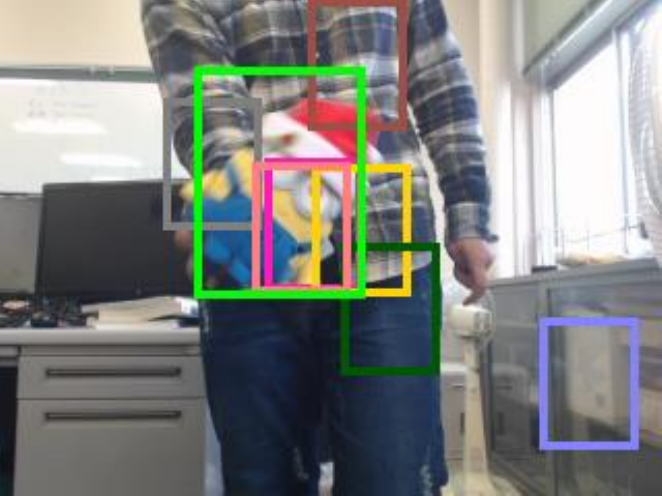}}
   \hspace{0.001in}
 \subfigure{
  \label{fig:subfig:e} 
  \includegraphics[width=0.05\linewidth]{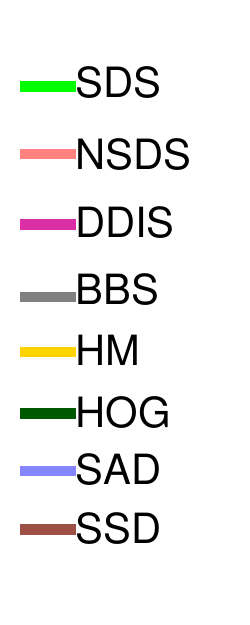}}
 \label{fig:subfig} 
 \caption{Scalable diversity similarity (SDS) for template matching. A doll moves from far to near in 3D space, mainly with its size, rotation angle and background changed in each frame. (a) Reference image. The template is marked by a red rectangle. (b), (c), (d) and (e) are the matching results over different frames with different methods. Our results are shown in green (SDS) and pink (NSDS). Best viewed in color.}
 \label{fig:head} 
\end{figure*}

Template matching is a basic component in a variety of computer vision applications. In this paper, we address the problem of template matching in unconstrained scenarios. That is, a rigid/nonrigid object moves in 3D space, with variant/invariant background and the object may undergo rigid/nonrigid deformations and partial occlusions, as demonstrated in \mbox{Figure. \ref{fig:head}}.

As the most crucial technique in template matching tasks, similarity measure has been studied for decades and yields in various methods from the classic sum of absolute differences(SAD), the sum of squared distances (SSD) to recent best buddies similarity (BBS) \cite{oron2018best} and deformable diversity similarity (DDIS) \cite{talmi2017template}. However, several aspects still need to be improved:
(1) Most real applications prefer showing matching results with bounding boxes in variable sizes to include object regions than a fixed size. Nevertheless, setting geometric parameters can result in an expansion of candidates for evaluation, which requires a distinctive similarity measure against scaling change.
(2) Template matching is usually dense and all the pixels/features within the template and a candidate window are taken into account to measure the similarity even some parts are not desirable (e.g., occlusions, appearance changes brought by deformation), this requires a similarity measure to deal with noises and outliers.
(3) Due to the possible deformation with the template, a good similarity measure is expected to be independent with the spatial correlation (e.g., when the object within a candidate window is strongly rotated compared to the template, the spatial correlation between the template and the candidate in raster scan order can become untrustworthy).
In this paper, scalable diversity similarity (SDS) is proposed to address the above problems. SDS can be applied with the multi-scale sliding window and is not limited by any specific parametric deformation models.

Both BBS and DDIS focus on settling the above problem (2) by exploiting the properties of the nearest neighbor (NN). Each NN is defined by a pair of patches between template and target. In BBS, if and only if each patch in a patch pair is the NN of the other, a match is defined and the number of such matches determines the BBS score. DDIS further improves the BBS by introducing relevant diversity of patch subsets between the target and template, which leads to the robustness of BBS against the occlusions and deformation. Although these methods can deal with deformation within a window to some extent, there are limitations especially on the problem (1) and (3). We extend DDIS to propose SDS based on the relevant diversity statistics. 

SDS has the following two advantages concerning the problem (1) and (3). The first is that SDS allows similarity measure between two sets of points in different sizes, and the magnitude of the score is scale-robust. Usually, the magnitude of the DDIS or BBS score grows with the increase of the point set's scale, which makes the larger candidate windows more favorable to be selected as final results. To alleviate the unfairness, SDS introduces bidirectional relevant diversity and penalizes on the change of scales to make the employment of multi-scale sliding window feasible, and the score can converge to the correct scale. This property of SDS is well statistically justified in Sec. \ref{sec:statistical}.

The second advantage of SDS is its robustness to the intense rotation. Both BBS and DDIS involve a spatial distance term in NN search or the final similarity calculation, which poses a limitation that the NN of a point must be spatially close. The limitation is a strong prior that can indeed reduce the number of outliers, but at the same time decrease the robustness against intense rotation. In this paper, instead of Cartesian coordinate, we exploit the polar angle of the polar coordinate for the calculation of spatial distance, which releases the limitation brought by the prior. Besides, rank information of appearance within a local circle is employed for searching NN along with local appearance, which helps to find more confident NN and yields in a significant improvement for intense rotation cases. This property of SDS is also statistically justified in Sec. \ref{sec:statistical}.

To summarize, the main contributions of this paper can be concluded as (a) SDS introduces bidirectional relevant diversity and penalizes on the change of scales to deal with scaling. (b) The rank of local appearance information and the polar radius is exploited to make the SDS robust against intense rotation change. (c) We originally collect a comprehensive dataset with 498 template-target pairs in the unconstrained environments for testing the matching performance, which includes 166 image pairs for scaling, rotation, scaling+rotation, respectively.
\subsection{Related work}
Template matching is a classic research topic mainly for object localization. The mechanism is straightforward: a large number of candidate windows are sampled in the target image, followed by a similarity measure between each candidate window and template. The similarity score plays a core role in measuring confidence and distinguishing the true target from the other candidates. Most widely used off-the-shelf measures are pixel-wise methods such as sum of difference (SSD), sum of absolute difference (SAD) and normalized cross-correlation (NCC), owing to their simplicity and efficiency. To deal with geometric changes on the target, extending the candidate sampling step with planar parametric transformations have been considered in many works, such as translation \cite{elboher2013asymmetric, chen2003fast, pele2008robust}, similarity transformation \cite{kim2007grayscale}, affine transformation \cite{korman2013fast, BMVC2015_121} and projective transformation \cite{zhang2016robust}. However, these methods usually fail in complex deformations because the pixel-wise similarity method relays on the correct correspondences between the pixels in template and candidate, which is highly limited by the planar geometric models.

In unconstrained environments, to deal with nonrigid transformations and other noises, involving global information instead of pixel-wise local information for designing a robust similarity is a key cue. Histogram matching (HM) \cite{hafner1995efficient,comaniciu2000real,perez2002color}, which mainly measure the similarity between two color histograms, is not restricted by geometric transformation. However, it is usually not a good choice when background clutter and occlusions appear within the windows. Earth mover's distance (EMD) \cite{rubner2000earth} is proposed to measure the similarity between two probability distributions. Furthermore, a more robust approach \cite{oron2015locally} is proposed by using spatial-appearance representation to measure the EMD. Tone mapping similarity measure \cite{hel2014matching} is proposed for handling noise, which is approximated by a piece-wise constant/linear function. Asymmetric correlation \cite{elboher2013asymmetric} is proposed to deal with both the noise and illumination changes. Other measures focus on improving the robustness against noise as proposed in M-estimator \cite{chen2003fast,sibiryakov2011fast} and Hamming-based distance \cite{shin2007fast,pele2008robust}. We refer the interested readers to a comprehensive survey \cite{ouyang2012performance}.

An eye-catching family of similarity measures in recent years is to explore a global statistic property over the two point sets. Bi-directional similarity (BDS) \cite{simakov2008summarizing} proposes that two point sets are considered similar if all points of one set are contained in the other, and vice versa. Best-buddies-similarity (BBS) \cite{dekel2015best,oron2018best} counts the two-side NNs as a similarity statistic. Deformable diversity similarity (DDIS) \cite{talmi2017template} measures the diversity of feature matches between the two sets and is reported to outperform BBS by revealing the ``deformation'' of the NN field. Despite the robustness of BBS and DDIS against the transformations within the search windows, scaling and rotation on the whole search windows have not been considered. In this paper, we propose a scaling and rotation independent similarity measure which leads to a significant improvement and allows multi-scale template matching in unconstrained environments.

\section{Methodology}
Given a template cropped from a reference image and a target image related by unknown geometric and photometric transformations, our purpose is to design a similarity measure, which can distinctively localize a region in the target image that includes the same object with the template by finding the maximum value. Each candidate region in the target image is defined by a rectangular window, and the candidate windows in the target image are generated in a multiple-scale sliding window fashion. Taking the template image $T = \left \{t_i \right \}_{i = 1}^n$ and a candidate window $Q = \left \{q_j \right \}_{j=1}^m$ from target image $\mathcal{Q}  = \left \{ q_l \right \}_{l = 1}^M$ as inputs, a SDS score in real number can be calculated, where the $t_i$ and $q_j$ represent non-overlapped patch from the template and a candidate window, respectively. $t_i$ and $q_j$ can also be treated as points when $T$ and $Q$ are treated as point sets. $Q \subseteq \mathcal{Q} $, and $m\leq M$.


Nearest neighbor has been shown to be a strong feature for designing similarity measure in some prior researches. To better address the difference, we first recall BBS \cite{dekel2015best} which counts the number of bidirectional NN matches between $T$ and $Q$:
\begin{equation}
\label{For:BBS}
BBS = c\left | \{ \exists t_i\in T,\exists q_j\in Q :\mathrm{NN}(t_i,Q) = q_j\wedge \mathrm{NN}(q_j,T) = t_i \} \right |,
\end{equation}
where $\mathrm{NN}(t_i,Q) = argmin_{q_j\in Q}\mathrm{d}(t_i,q_j)$ is a function returns the NN of $t_i$ with respect to $Q$, and the $\mathrm{d(\cdot)}$ is a distance function. The $\left | \{ \cdot \} \right |$ denotes the size of a set, and the $c = 1/\mathrm{min}\{n,m\}$ is a normalization factor. 

We are now ready to introduce our method in a bottom-up fashion: from NN search to bidirectional diversity, and finally the SDS similarity.

\subsection{Rank of Local Appearance for Rotation Robust NN search}
The distance function in Eq. \ref{For:BBS} is defined by
\begin{equation}
\mathrm{d}\left(p_{i}, q_{j}\right)=\left\|p_{i}^{(A)}-q_{j}^{(A)}\right\|_{2}^{2}+\lambda\left\|p_{i}^{(L)}-q_{j}^{(L)}\right\|_{2}^{2},
\end{equation}
where $(A)$ denotes pixel appearance (e.g., RGB)
and $(L)$ denotes pixel location $(x, y)$ within the
patch normalized to the range [0, 1]. In the stage of NN searching, under the assumption that intense deformation such as rotation do not occur within the patch, the spatial term can contribute to improving the confidence of NN by confirming the consistency of appearance and position. We propose 
\begin{equation}
\label{distancerank}
\mathrm{d}\left(p_{i}, q_{j}\right)=\left\|p_{i}^{(A)}-q_{j}^{(A)}\right\|_{2}^{2}+\lambda\left\|p_{i}^{(R)}-q_{j}^{(R)}\right\|_{2}^{2},
\end{equation}
to incorporate $(R)$ instead of $(L)$, which denotes the rank with respect to the appearance of pixels within a circle. The origin of the circle is $p_i$, with a support radius of $r$. Specifically,
\begin{equation}
\label{For:rank}
    p_i^{(R)} = \sum_{p \in \mathrm{circle}(p_i,r)}\mathrm{I}\left (p_i^{(A)}\geq p^{(A)}\right)/r^2,
\end{equation}
where $\mathrm{I}(\cdot)$ is an indicator function that turns true and false into 1 and 0. Equation in the same form is applied to $q_j$. Unlike pixel location, 
the appearance rank defined by Eq. \ref{For:rank} is invariant to rotation, which can also be considered as structural information (e.g., the shape of the distribution of pixel values) extracted from a local region. As the rotation will not destroy the structure, it is reasonable to explain its invariance against rotation. Furthermore, the Euclidean distance of orders emphasizes the influence of local extremes, which also contributes to keeping the local features well.

\subsection{Bidirectional Diversity for Discriminative Similarity Measure}
\label{sec:bidirectional}
We first extend the diversity similarity (DIS) defined in \cite{talmi2017template} to a bidirectional way. The DIS is defined as 
\begin{equation}
\label{For:DIS}
DIS = c\left | \{ t_i\in T: \exists q_j\in Q , \mathrm{NN}(q_j,T) = t_i \} \right |,
\end{equation}
which counts the types of points in $T$ that have NN in $Q$ with the same pixel type (i.e., defined as diversity in direction $T \rightarrow Q$). The authors claim that this one direction diversity provides a good approximation to BBS with less computation. However, the number of candidates increase explosively by allowing multi-scale candidate windows $Q$, therefore a more discriminative similarity measure is needed. We exploit both diversity calculated with respect to $T$ and $Q$ (i.e., $T \rightarrow Q$ and $Q \rightarrow T$). Specifically, we first define the following function $\varepsilon(t_i)$ which indicates the number of points $q_j\in Q$ whose NNs are equal to $t_i$ in direction $T \rightarrow Q$,
\begin{equation}
\label{For:varepsilon}
\varepsilon (t_i) = \left | \left \{ q_j\in Q: \mathrm{NN}(q_j,T)=t_i\right \}\right |,
\end{equation}
%
where NN($\cdot$) here is calculated with distance defined in Eq. \ref{distancerank}. To understand the equation, we analyze its relationship with diversity from two situations. For $\vert T \vert=\vert Q \vert$: (1) When $\varepsilon(t_i)\geq 1$, the value is inversely proportional to the diversity contribution. That is, large value of $\varepsilon{(t_i)}$ indicates that many points in $Q$ have the same NN of $t_i$, which will lower the diversity defined in Eq. \ref{For:DIS}. (2) When $\varepsilon (t_i)= 0$, it indicates that a $t_i$ is not a NN of any $q_i$, which also hinders the increase of diversity. An ideal situation is that for each $t_i$, $\varepsilon (t_i)=1$. For $s\vert T \vert = \vert Q \vert$, the situations become more complex. (1) when $ \varepsilon(t_i)= 0$, similarly it means low contribution to the diversity. (2) Due to the scaling $s$ between $Q$ and $T$, one point can be the NN of multiple points, when $1 \leq \varepsilon(t_i) \leq s$, it contributes to the diversity. (3) When $\varepsilon (t_i)>s$, it will lower the diversity.


We propose to simultaneously introduce this statistic to direction $Q \rightarrow T$. However, it is not straightforward in the case of template matching. Because the candidate window $Q$ usually belongs to a target image $\mathcal{Q}$, where $\vert \mathcal{Q} \vert \gg \vert Q \vert$. That is, when finding NNs in the case of $T \rightarrow Q$, as $T$ is fixed and the preprocessing (e.g., sorting for brute force search, building kd-tree, etc.) only need to be conducted once. In the case of $Q \rightarrow T$, as such preprocessing for NN search has to be conducted over each $Q$, it will suffer from time cost. To tackle this problem, we pose an assumption that $\mathrm{NN}(t_i,Q)$ has a high probability to be included in the set of $k$ approximate NNs with respect to $\mathcal{Q}$, which is denoted by $\mathrm{ANN}(t_i,\mathcal{Q})$. Formally, we define the following function which counts the number of points (i.e., patches in the image) $t_i\in T$ whose ANNs include $q_j$ in direction $Q \rightarrow T$,
\begin{equation}
\label{For:tau}
\tau(q_j) = \left | \left \{t_i\in T, Q \in \mathcal{Q}: q_j \in \mathrm{ANN}^k(t_i,\mathcal{Q})\right \}\right | .
\end{equation}
\subsection{Scalable Diversity Similarity}
With bidirectional diversity $\varepsilon(t_i)$ and $\tau(q_j)$ defined, we define the SDS to quantify the the similarity between template $T$ and candidate $Q$ with given target image $\mathcal{Q}$ and scaling $s$ as follows, where $s$ can be calculated from $T$ and $Q$,

\begin{equation}
\label{For:SDS}
\mathop{SDS (T, Q, s, \mathcal{Q})} = \lambda_1 \frac{\sum_{q_j} \mathrm{I}(\tau(q_j) \neq 0) \sum_{t_i} \mathrm{I}(\varepsilon(t_i) \neq 0)}{ \sum_{q_j}\vert \rho(q_j) - s\rho(NN(q_j,T))\vert}U.
\end{equation}
Where parameter $\lambda_1$ is a normalization factor inversely proportional to the increase of $s$ (e.g., $\lambda_1=s^{-1}$). As analyzed in Sec. \ref{sec:bidirectional}, only points in $T$ which hold $\varepsilon(t_i)\neq 0$, and points in $Q$ which hold $\tau(q_j)\neq 0$ can possibly contribute to the increase of the diversity. $\rho(\cdot)$ returns the radius of a pixel in polor coordinate, with the pole set as the according geometric center of $T$ and $Q$. The denominator of Eq. \ref{For:SDS} penalizes the spatial consistency in polar coordinate, to further increase the robustness against in-plan rotation. Term $U$ is a normalization term for the number of NNs with respect to scaling. Following the analysis in Sec. \ref{sec:bidirectional}, in our implementation, $U$ is defined as $\sum_{t_i,\varepsilon (t_i)> 0}\exp\left ( \mathrm{I}( s/ \varepsilon (t_i)\geq 1)+\mathrm{I}(s/ \varepsilon (t_i)<1)s/ \varepsilon (t_i) -1 \right )$, which increases when more $t_i$ holds $s/ \varepsilon (t_i) \geq 1$. In conclusion, SDS can be viewed as a cooperation of three terms: (1) The numerator term to evaluate the bidirectional diversity, (2) the denominator term to evaluate the spatial consistency, (3) the $U$ term to normalize the number of NNs with respect to $s$ .
\begin{figure*}[t]
 \centering
  \subfigure[SSD, $s=1$]{
  \label{fig:subfig:a} 
  \includegraphics[width=0.15\linewidth]{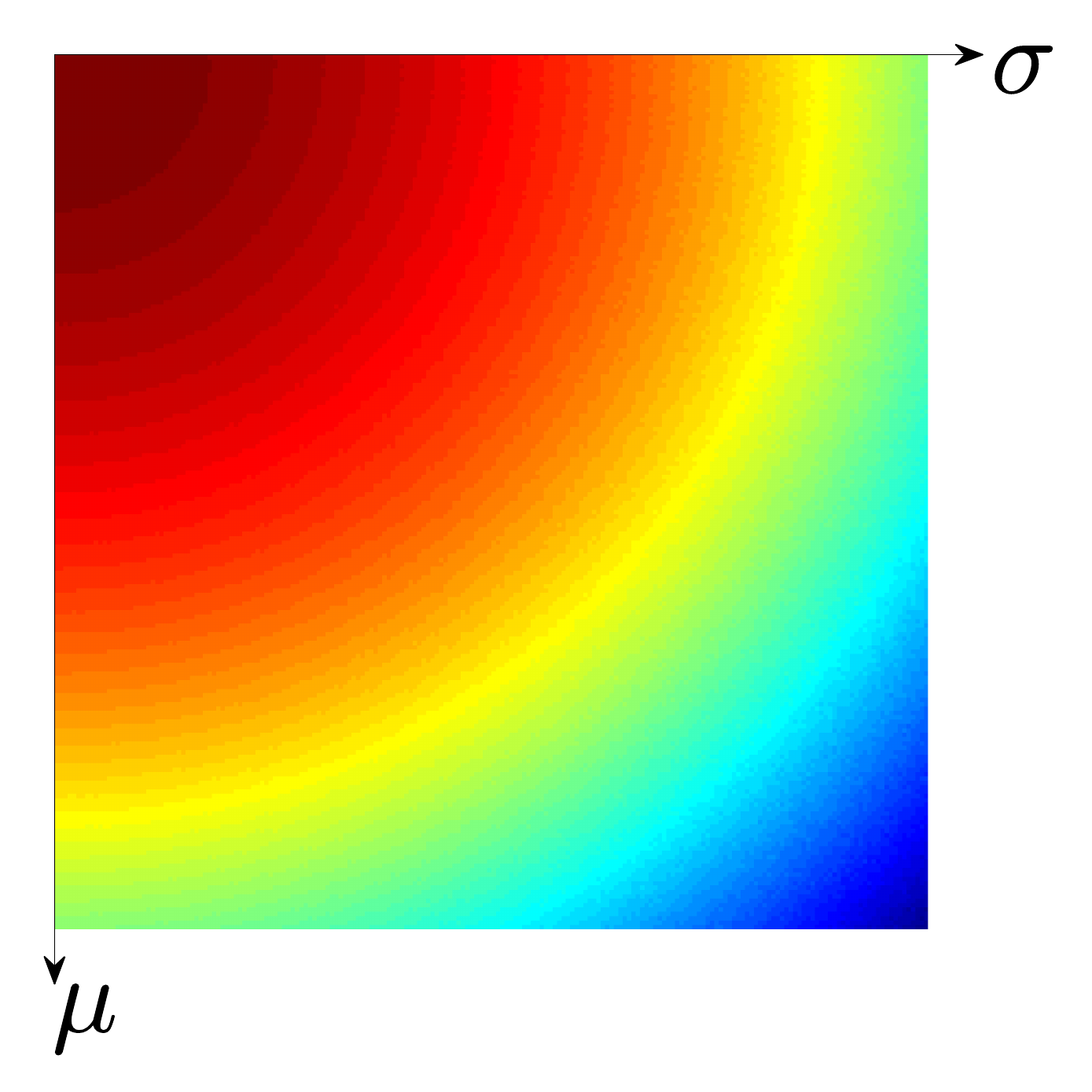}}
 \subfigure[BBS, $s=1$]{
  \label{fig:subfig:b} 
  \includegraphics[width=0.15\linewidth]{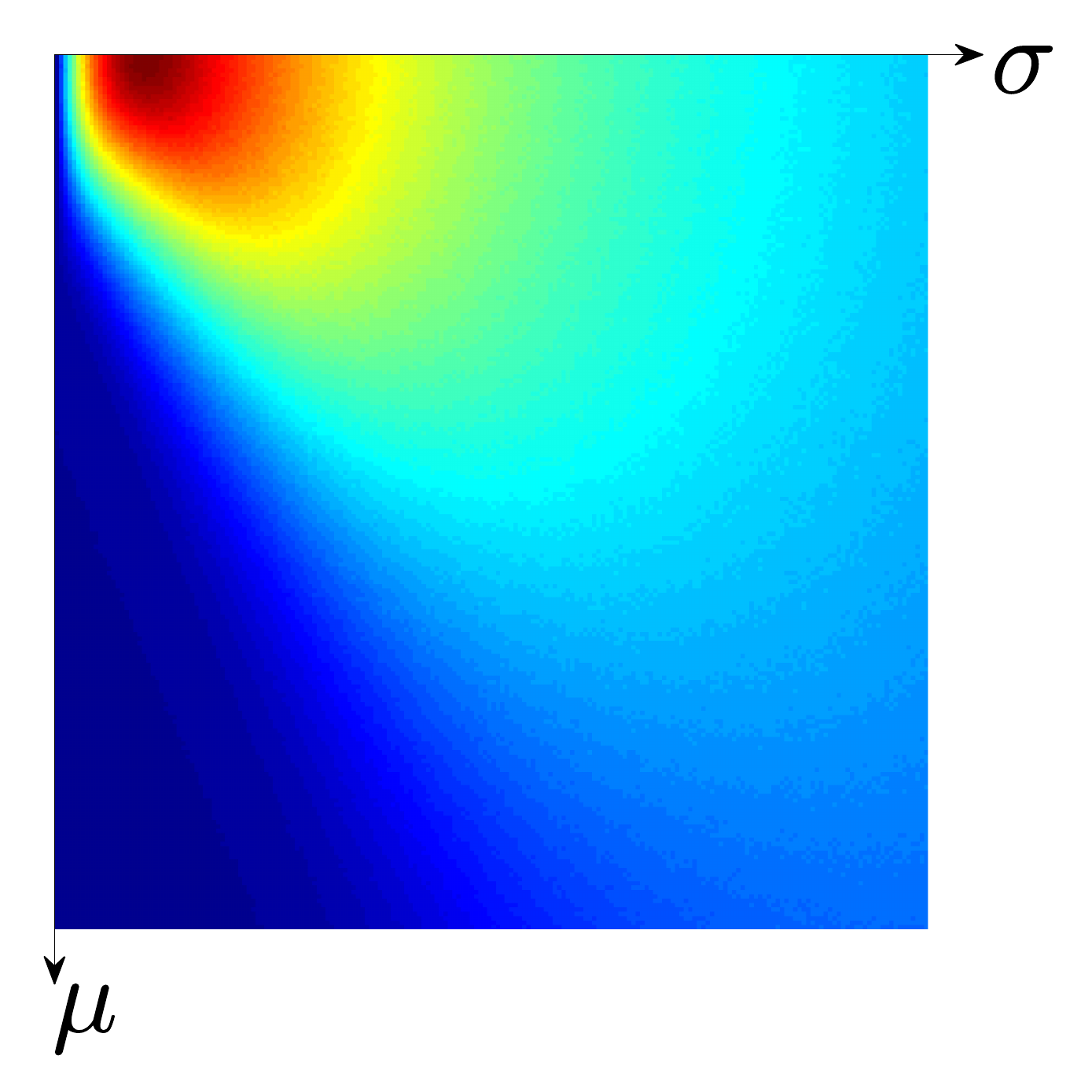}}
  \subfigure[DDIS, $s=1$]{
  \label{fig:subfig:b} 
  \includegraphics[width=0.15\linewidth]{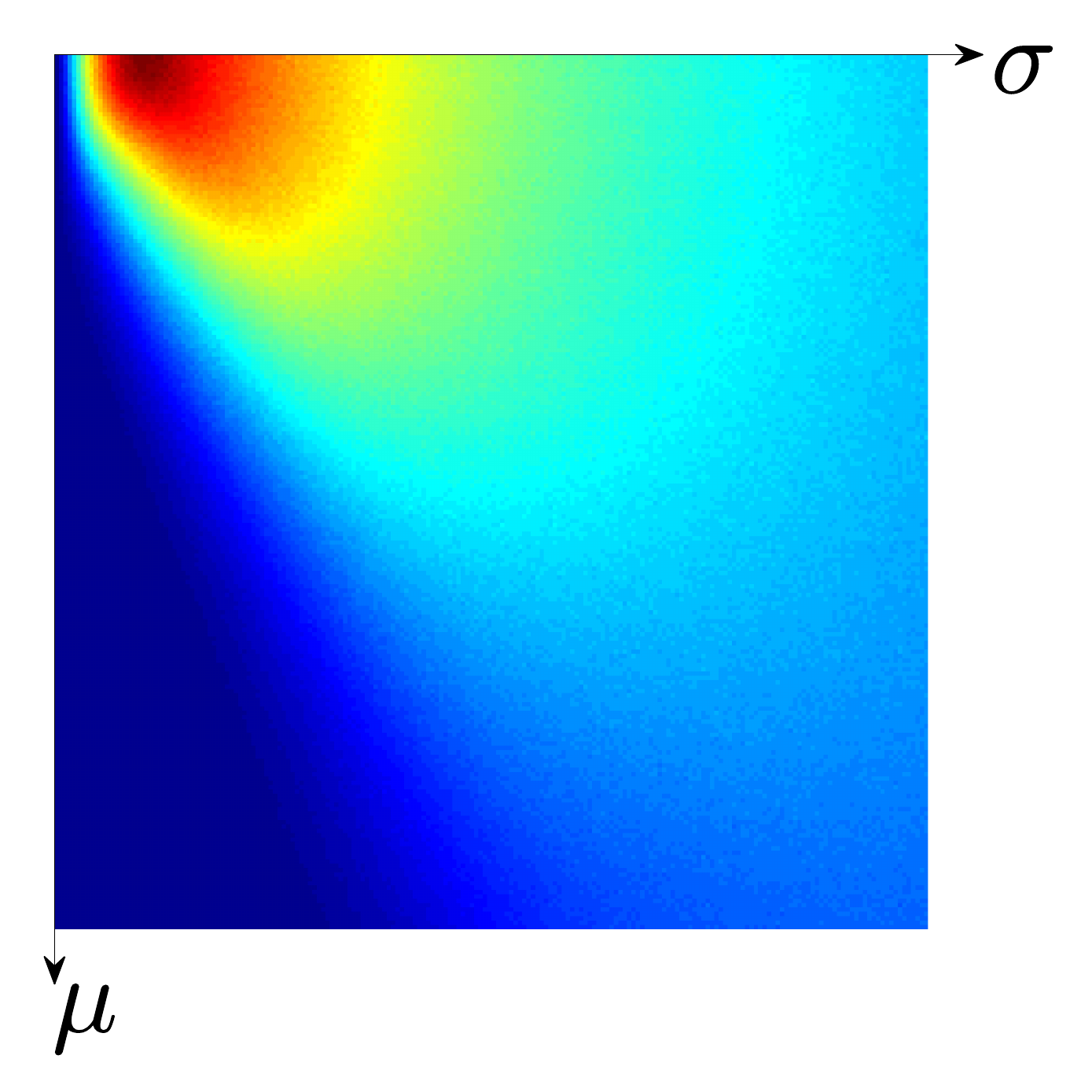}}
 \subfigure[SDS, $s=1$]{
  \label{fig:subfig:c} 
  \includegraphics[width=0.15\linewidth]{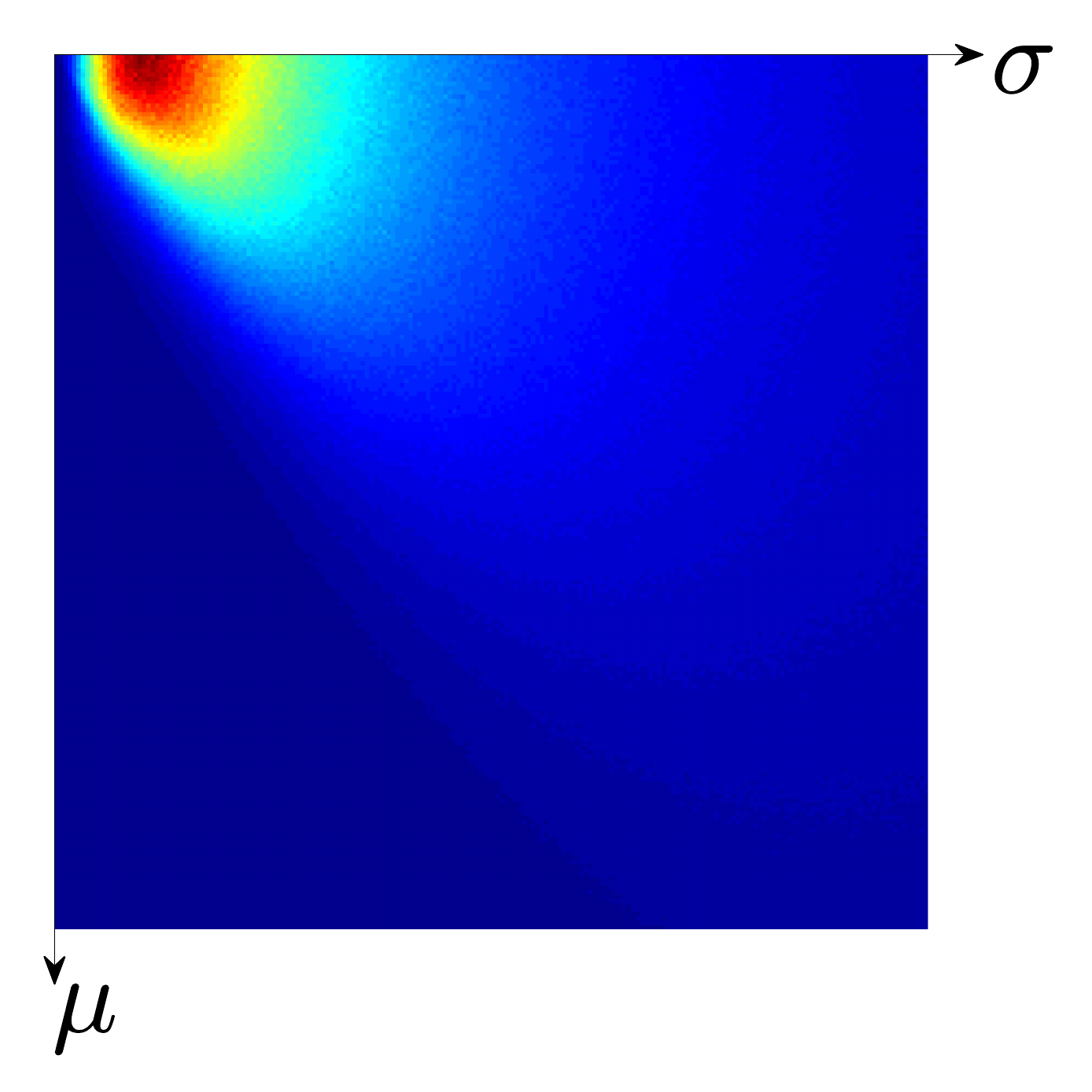}}
 \subfigure[SDS, $s=0.5$]{
  \label{fig:subfig:d} 
  \includegraphics[width=0.15\linewidth]{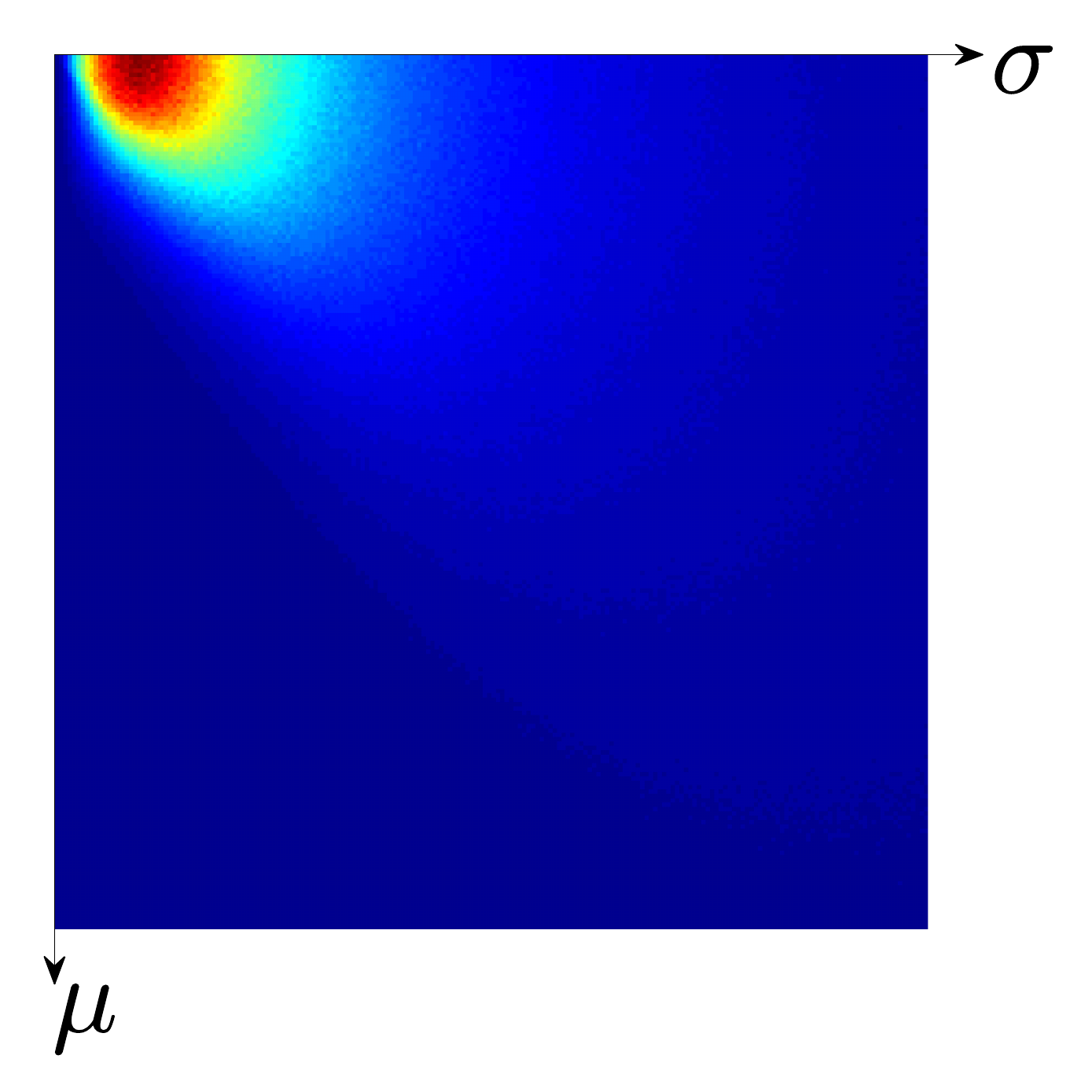}}
 \subfigure[SDS, $s=2$]{
  \label{fig:subfig:e} 
  \includegraphics[width=0.15\linewidth]{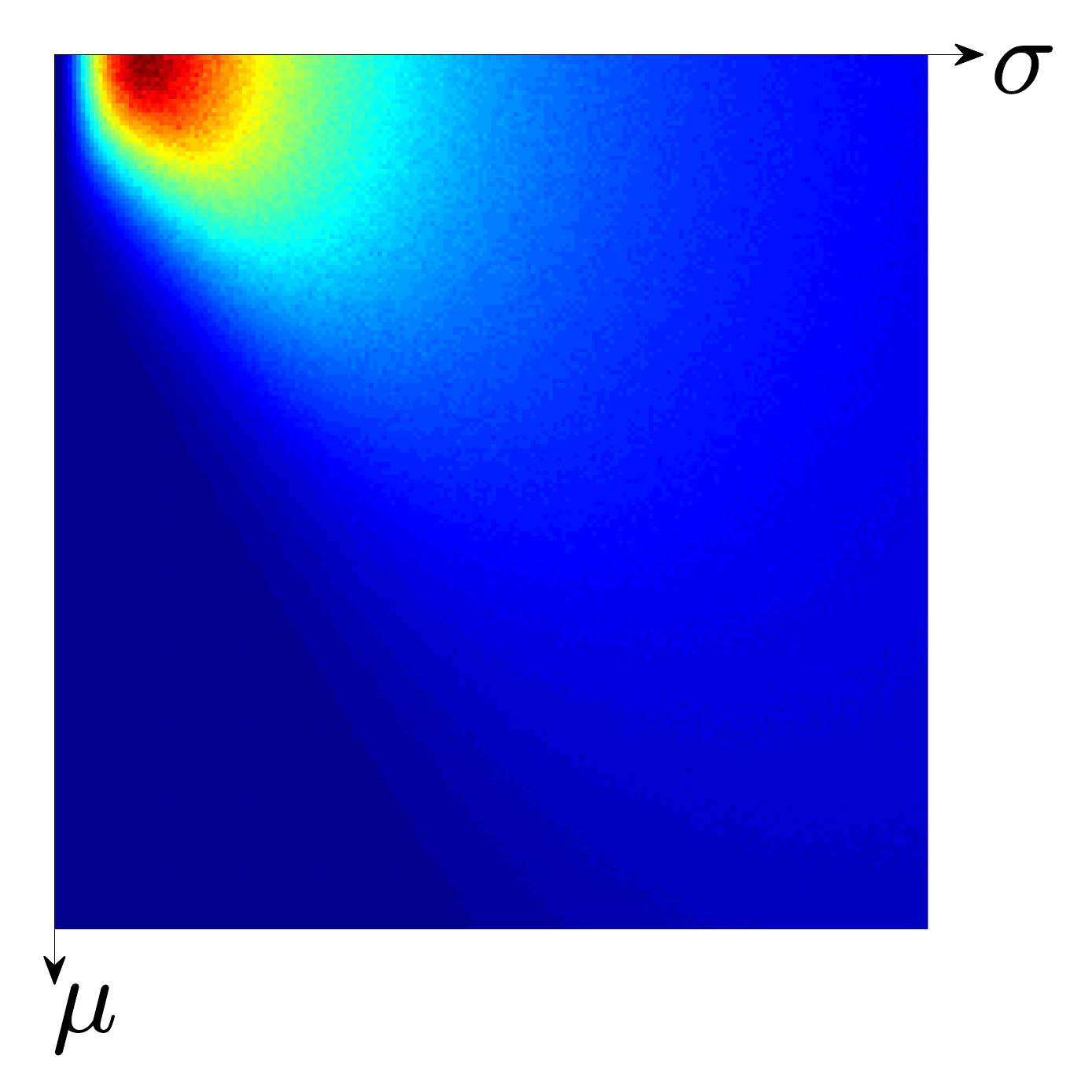}}
 \label{fig:subfig} 
 \caption{Expectation maps of SSD, BBS, DDIS, and SDS in 1D Gaussian case. Two points sets, $T$ and $Q$ are randomly drawn from two 1D Gaussian models $N(0,1)$ and $N(\mu,\sigma)$, respectively. $\mathcal{Q}$ is set to be the same with $Q$. In (a), (b), (c) and (d), $\vert T \vert$ and $\vert Q \vert$ are set to 100 (i.e., with a fixed scale). In (e), $\vert T \vert=100$ and $\vert Q \vert=50$ (i.e., $s=0.5$). In (f), $\vert T \vert=100$ and $\vert Q \vert=200$ (i.e., $s=2$). The parameters of the Gaussian for generating $Q$ increase from left-top $(\mu=0,\sigma=0)$ to right-bottom. It can be clearly observed that SDS drops fastest when $(\mu \neq 0,\sigma \neq 1)$, and remains well against scale change.}
 \label{fig:expectationgraph} 
\end{figure*}
\subsection{Statistical Analysis}
\label{sec:statistical}

 \begin{figure*}[t]
 \centering
  \subfigure{
  \label{fig:Convergent:b} 
\includegraphics[width=0.024\linewidth]{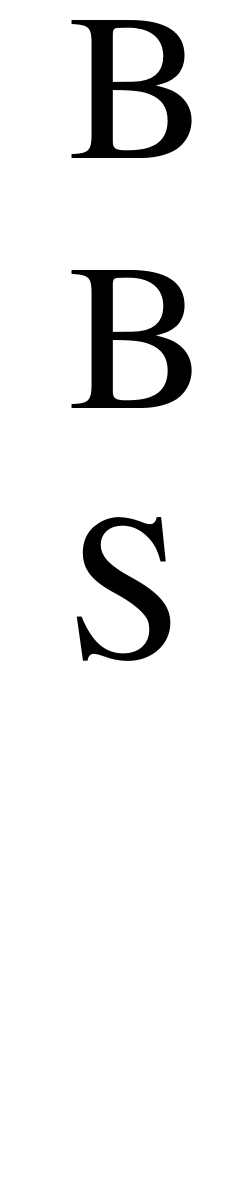}}
\hspace{0.001in}
    \subfigure{
  \label{fig:Convergent:a} 
  \includegraphics[width=0.17\linewidth]{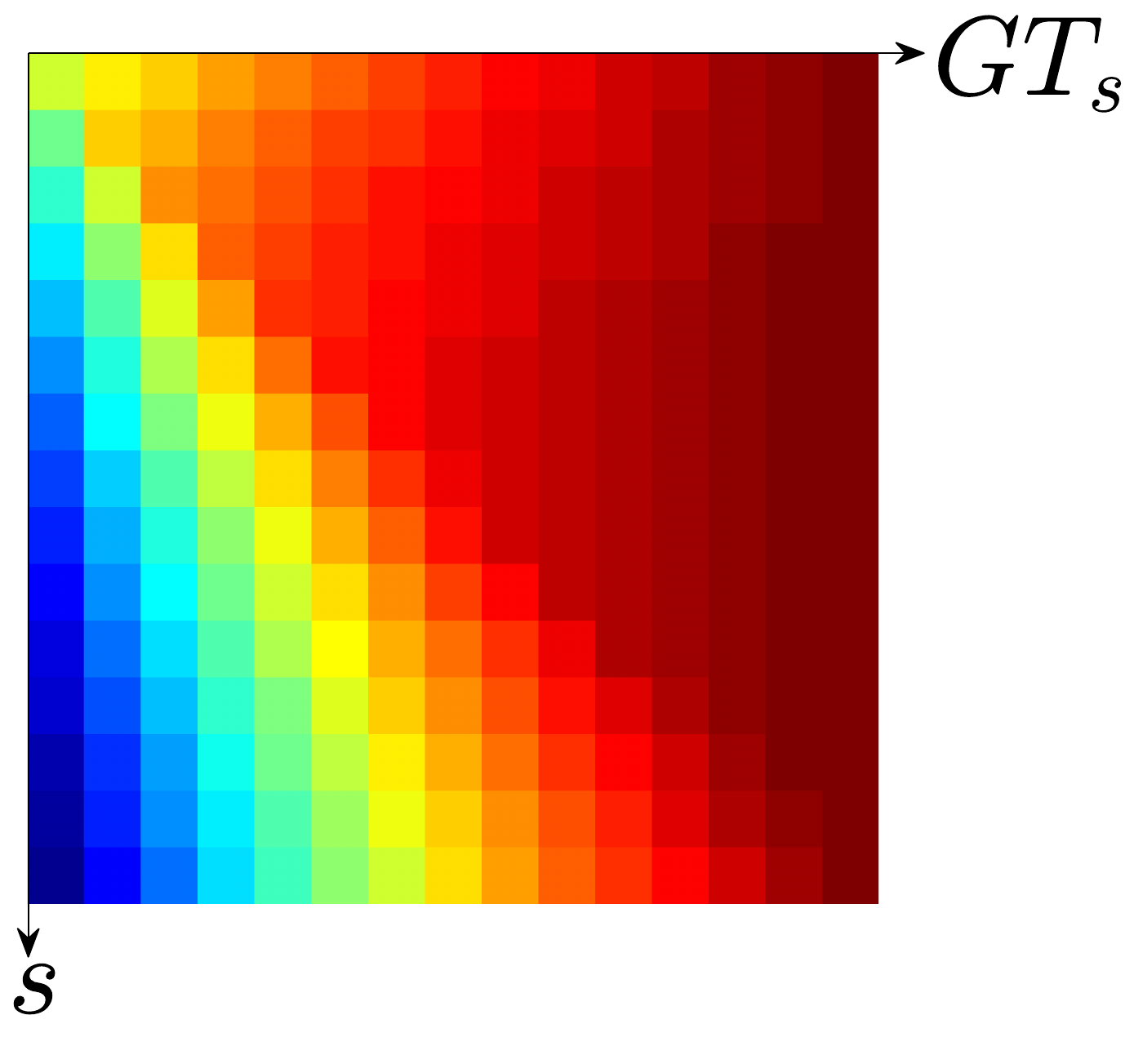}}
 \subfigure{
  \label{fig:Convergent:b} 
  \includegraphics[width=0.22\linewidth]{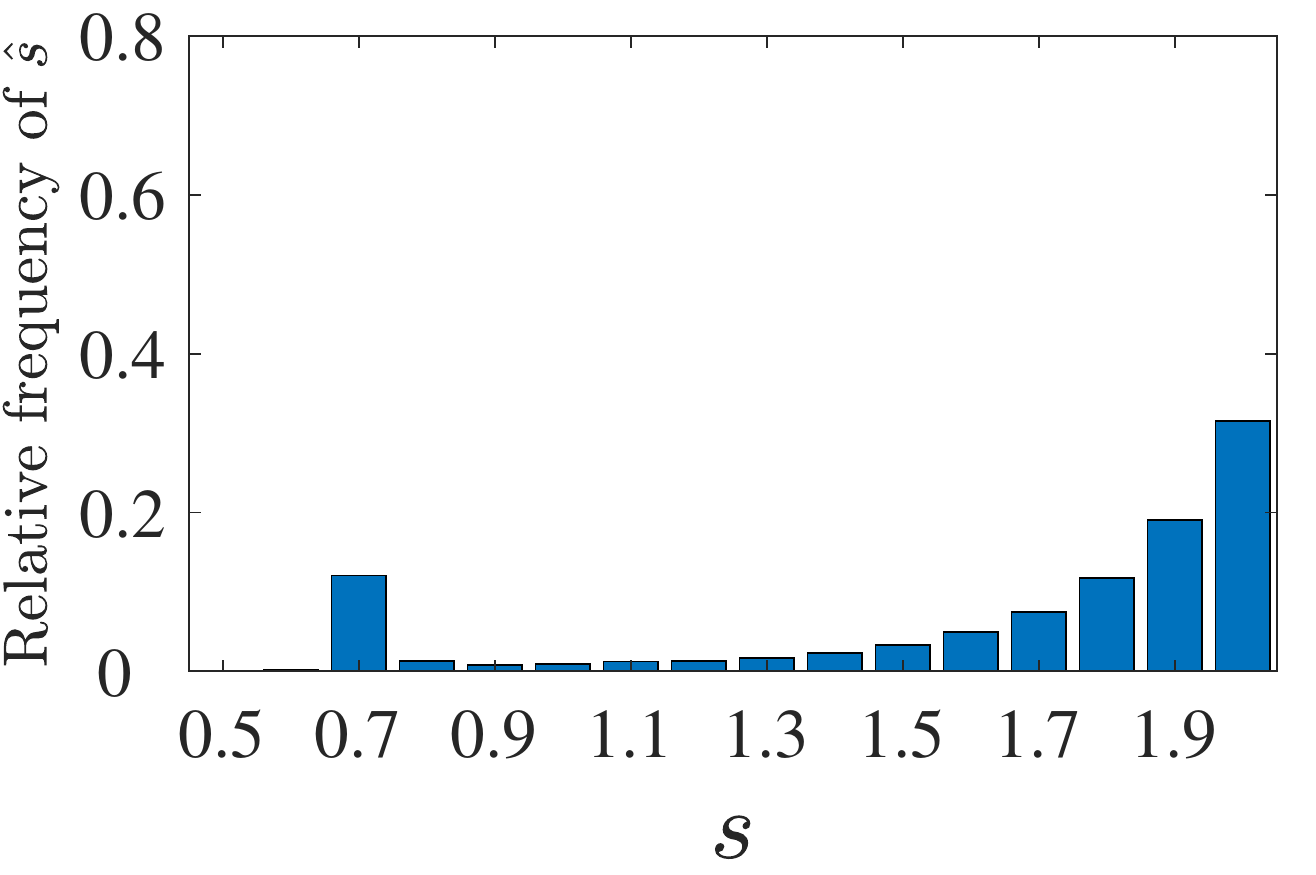}}
 \subfigure{
  \label{fig:Convergent:c} 
  \includegraphics[width=0.22\linewidth]{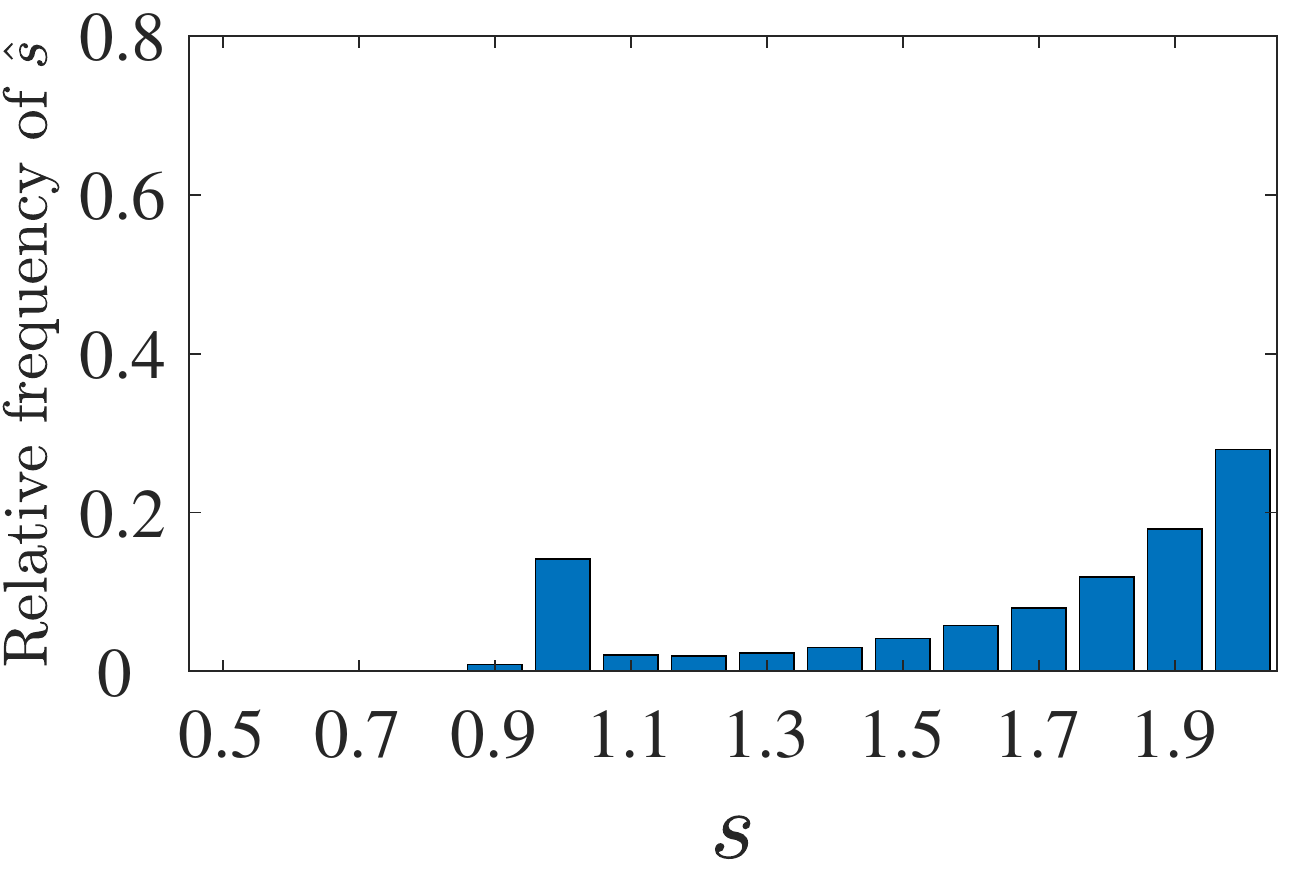}}
 \subfigure{
  \label{fig:Convergent:d} 
  \includegraphics[width=0.22\linewidth]{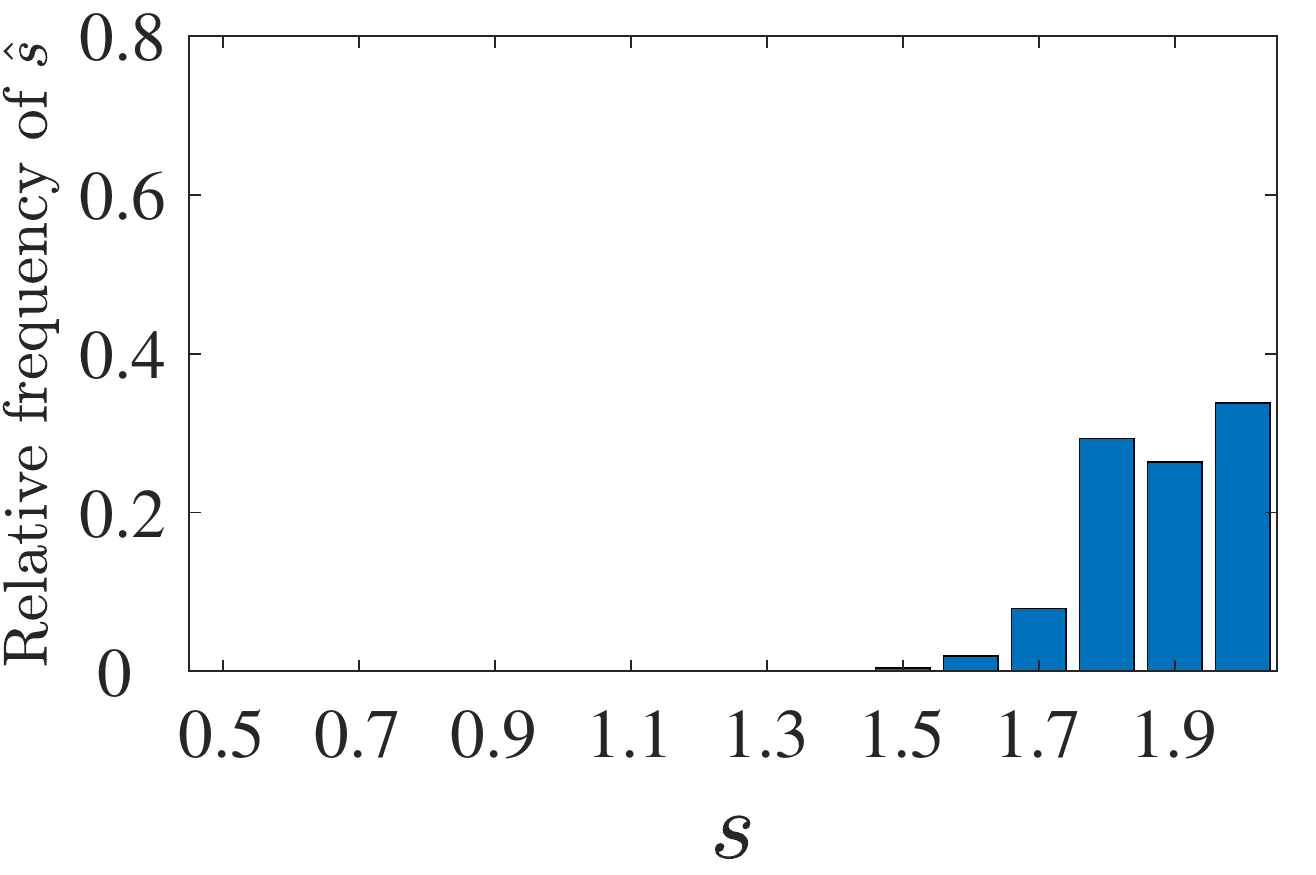}}
  
    \subfigure{
  \label{fig:Convergent:b} 
\includegraphics[width=0.024\linewidth]{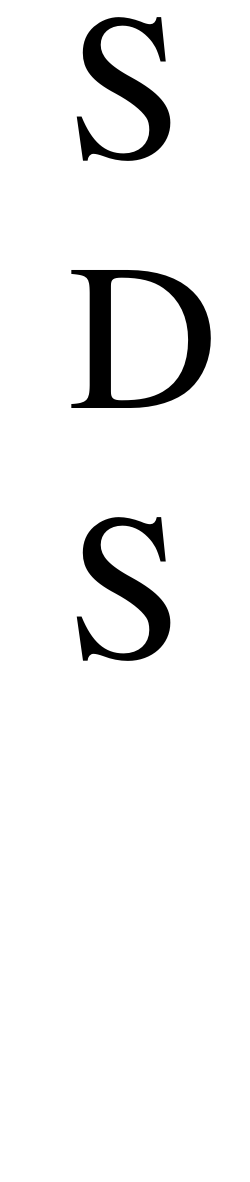}}
  \setcounter{subfigure}{0}
    \subfigure[Expectation]{
  \label{fig:Convergent:a} 
  \includegraphics[width=0.17\linewidth]{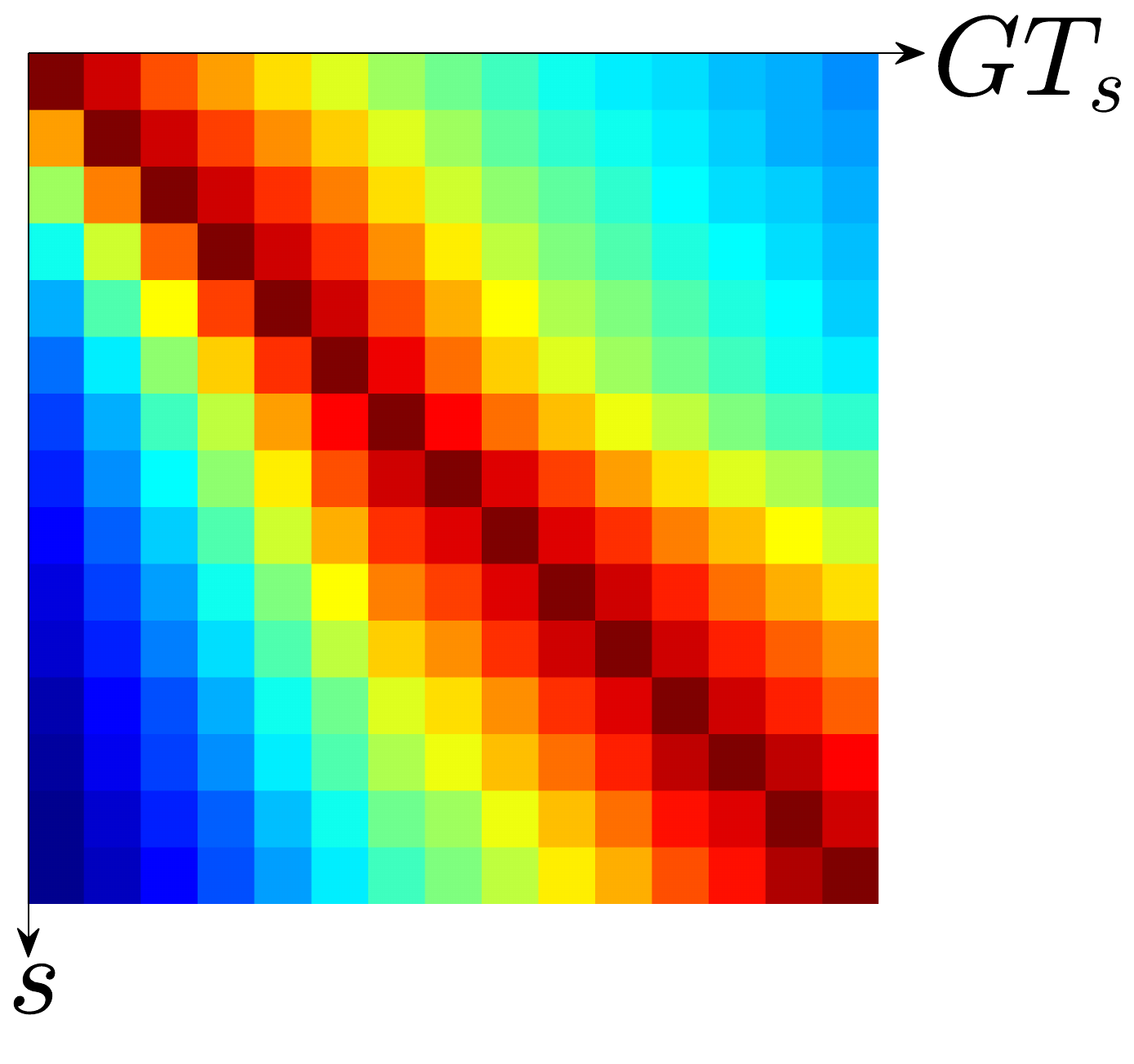}}
 \subfigure[Ground truth s=0.7]{
  \label{fig:Convergent:b} 
  \includegraphics[width=0.22\linewidth]{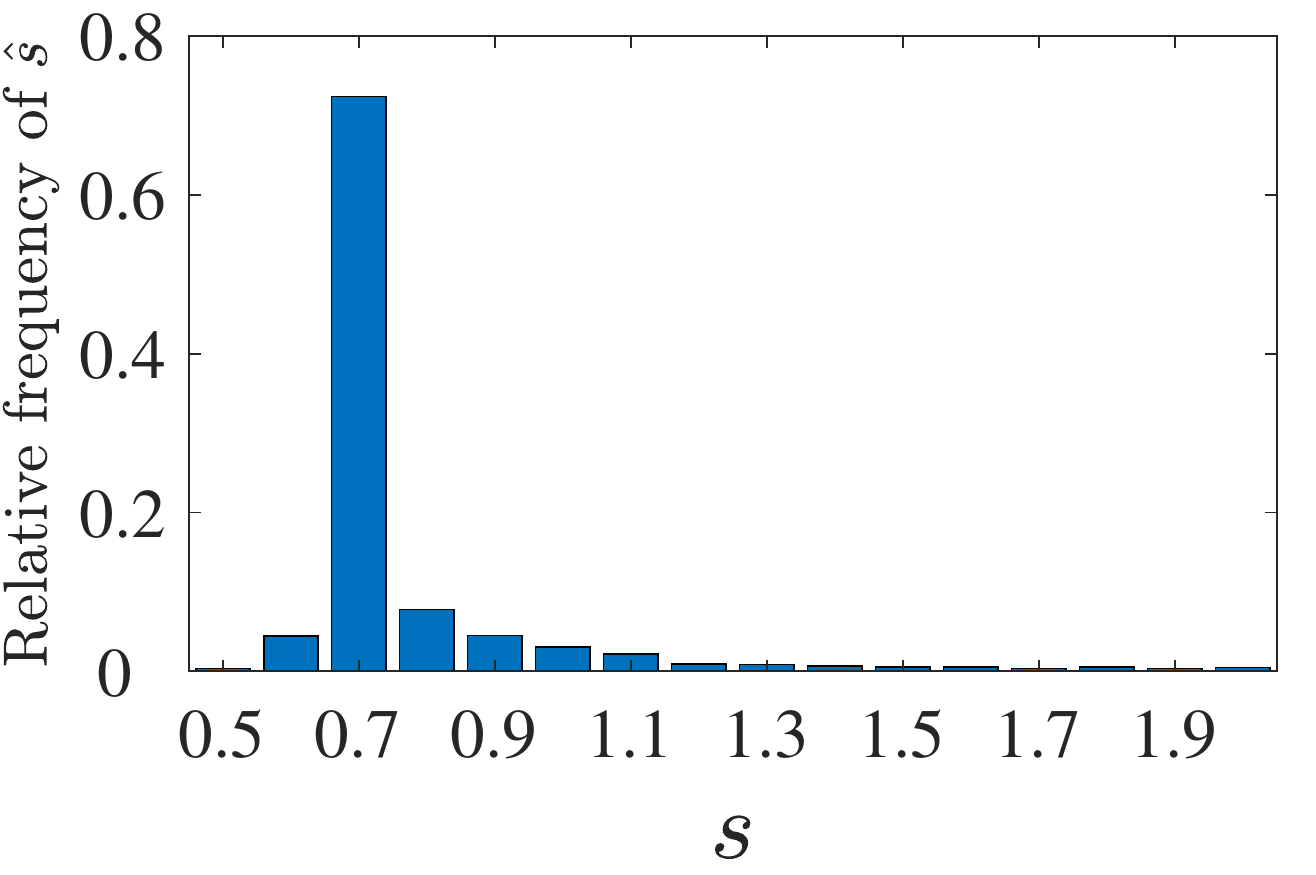}}
 \subfigure[Ground truth s=1.0]{
  \label{fig:Convergent:c} 
  \includegraphics[width=0.22\linewidth]{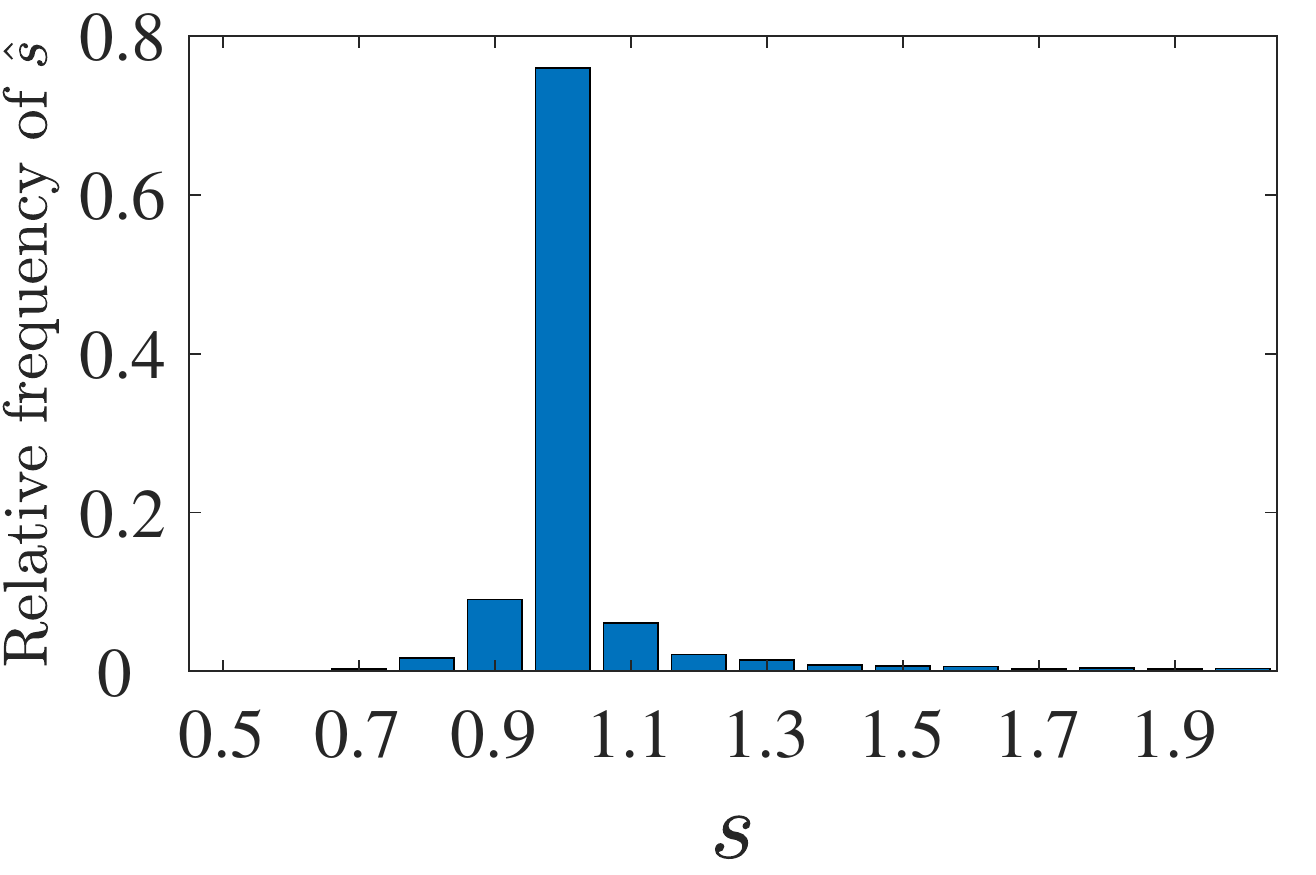}}
 \subfigure[Ground truth s=1.8]{
  \label{fig:Convergent:d} 
  \includegraphics[width=0.22\linewidth]{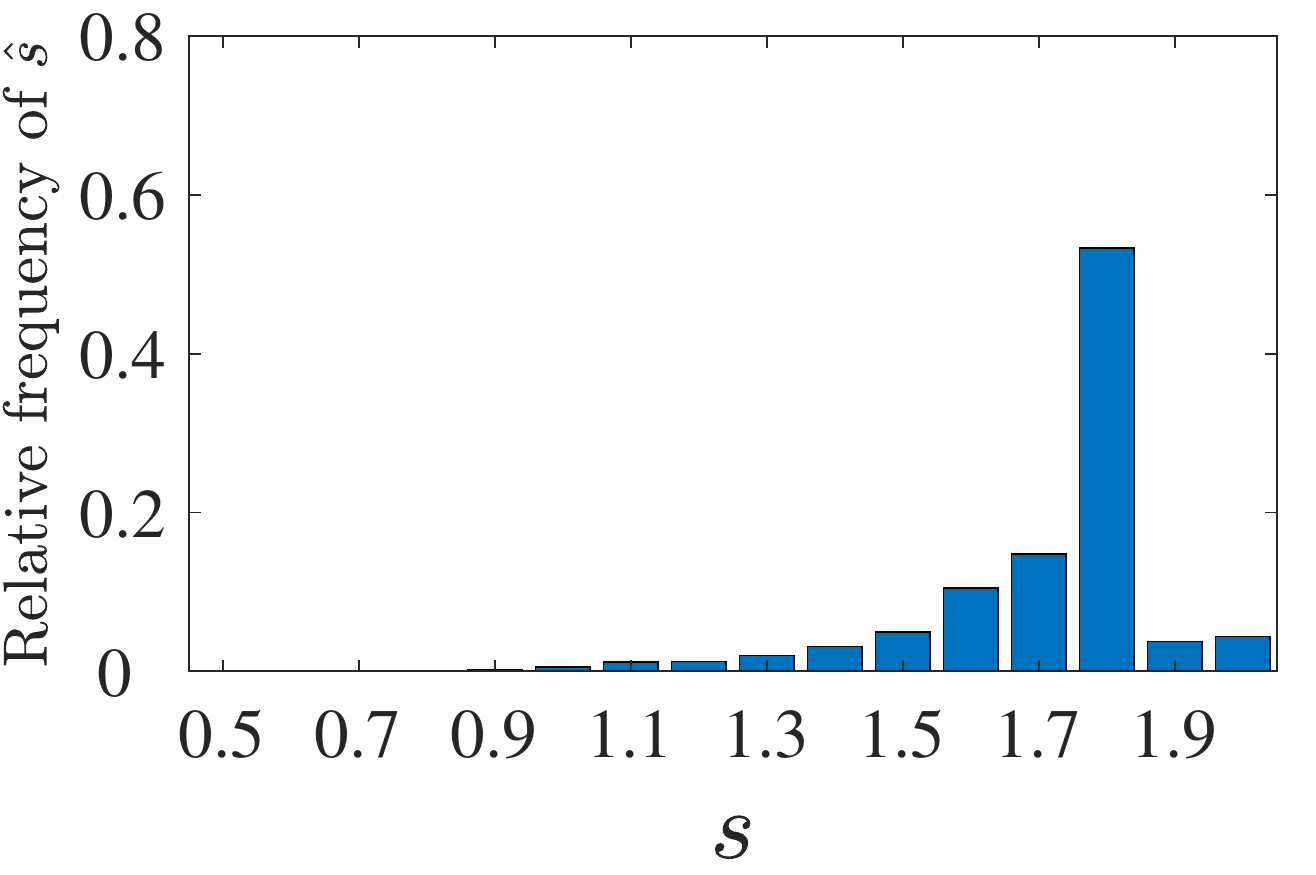}}

 \caption{Scale estimation by similarity maximization. (a) shows the approximated expectation map with respect to the variation of ground truth $GT_s$ and $s$. SDS (bottom row) achieves maximum expectation values on the diagonal while BBS (top row) fails in estimating the proper scale. (b)$\sim$(d) demonstrate the normalized histogram of estimated $\hat{s}$ based on random trials. In the case of SDS (bottom row), the according bin of $s=GT_s$ achieves highest frequency. BBS (top row) performs well in a local scale range while fails in the global.}
 
  \label{fig:Convergent}
\end{figure*}
\textbf{Analysis of scaling-robustness.} To assert the effectiveness of SDS in measuring the similarity between scale-variant point sets, we first provide a 1D statistical analysis following \cite{dekel2015best,talmi2017template}. The expectations of similarity between two point sets drawn from two 1D Gaussian models are calculated for comparison, where point sets are cast as template/candidate window, points are cast as patches. Our goal is to show that the expectation of SDS is maximal when the two Gaussian models are the same and decrease fastest when models separate. We further analyze the expectations of point sets in different scaling size to show the scaling-robustness of SDS. As suggested by \cite{talmi2017template}, Monte-Carlo integration is exploited for approximating the expectation. Figure \ref{fig:expectationgraph} (a) to Figure. \ref{fig:expectationgraph} (d) show the illustration of approximated expectation maps when two point sets have same size ($s=1$). It can be obviously observed that the expectation of SDS drops faster than either SSD, BBS, or DDIS when the parameters of the second Gaussian ($\mu$ and $\sigma$) get away from the parameters of the first Gaussian ($\mu=0$ and $\sigma=1$). Figure \ref{fig:expectationgraph} (d) to Figure. \ref{fig:expectationgraph} (f) show the comparison of expectation map when two point sets are in different sizes ($s=1, 0.5, 2$), which provides a strong evidence that SDS is highly robust against scaling as the expectation maps almost remain the same despite the scaling change.

To further show that the scale of target with respect to $T$ can be estimated by maximizing SDS, we provide a statistical result in Figure. \ref{fig:Convergent}. Similar with Figure. \ref{fig:expectationgraph}, $T$ is drawn from $N(0,1)$ and $Q$ is generated for expectation approximation. The difference is, we further prepare $\mathcal{Q}$ which involves background points to simulate the template matching task. Here, $\mathcal{Q}=T\cup B$, $GT_s\vert T \vert + \vert B \vert = \vert \mathcal{Q} \vert$ and $B$ is composed of background points drawn from $N(\mu,\sigma)$, with $\mu \in [0,10], \sigma \in [0,10]$. In this demonstration, $\vert T \vert$ and $\vert \mathcal{Q} \vert$ are set to 100 and 200 respectively. $\vert Q \vert=s\vert T \vert$ and $s$ varies from 0.5 to 2 with step of 0.1. The $Q$ can be treated as a candidate window in the template matching task and is sampled from $\mathcal{Q}$ by preferentially sample points in $T$ (i.e., nearest neighbor interpolation). For example, when $s=1.5$, 150 points need to be sampled to formulate $Q$, with 100 points from $T$ and 50 points from $B$. Estimated $\hat{s}=\argmax_{s} SDS(T,Q,s,\cal{Q})$ is supposed to approximate the ground truth scale $GT_s$ well. This statistical analysis clearly prove the robustness of SDS against scaling, and the ability for estimating proper scale of the target.



 \begin{figure*}[t]
 \centering
  \subfigure[Example of $T$]{
  \label{fig:rotation:c} 
  \includegraphics[width=0.22\linewidth]{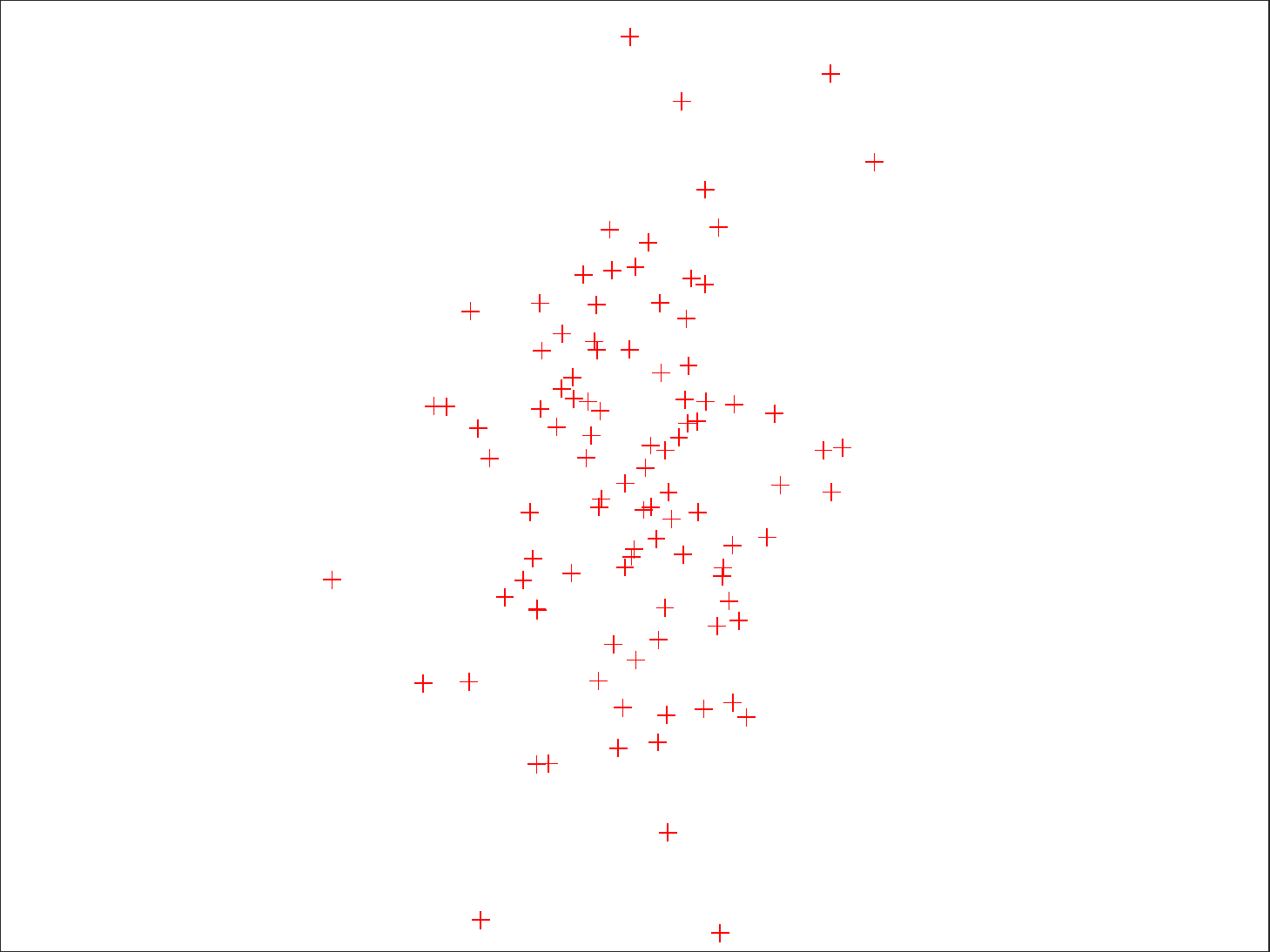}}
 \subfigure[Example of $Q$]{
  \label{fig:rotation:d} 
  \includegraphics[width=0.22\linewidth]{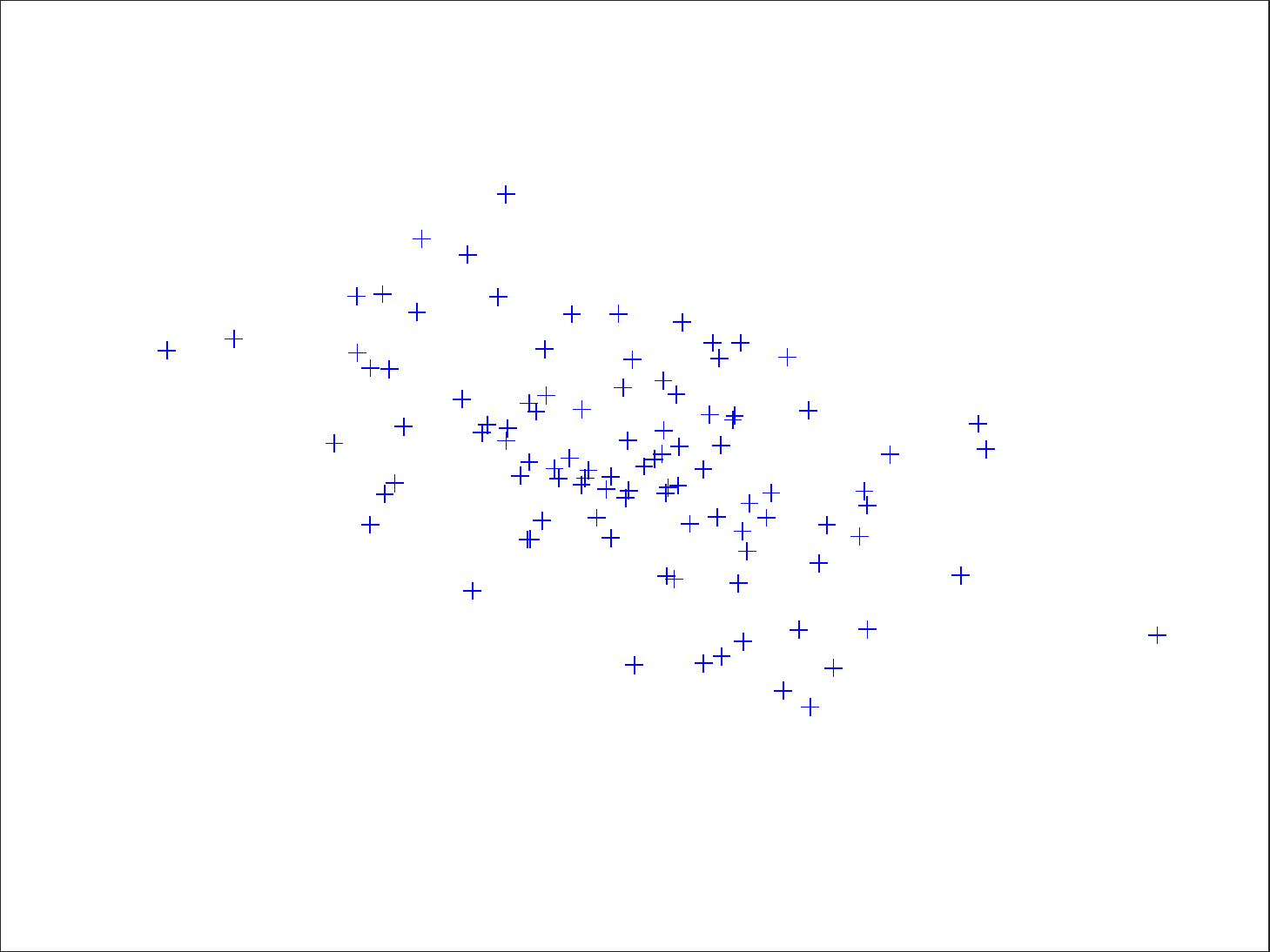}}
   \subfigure[$E(BBS)$]{
  \label{fig:rotation:b} 
  \includegraphics[width=0.22\linewidth]{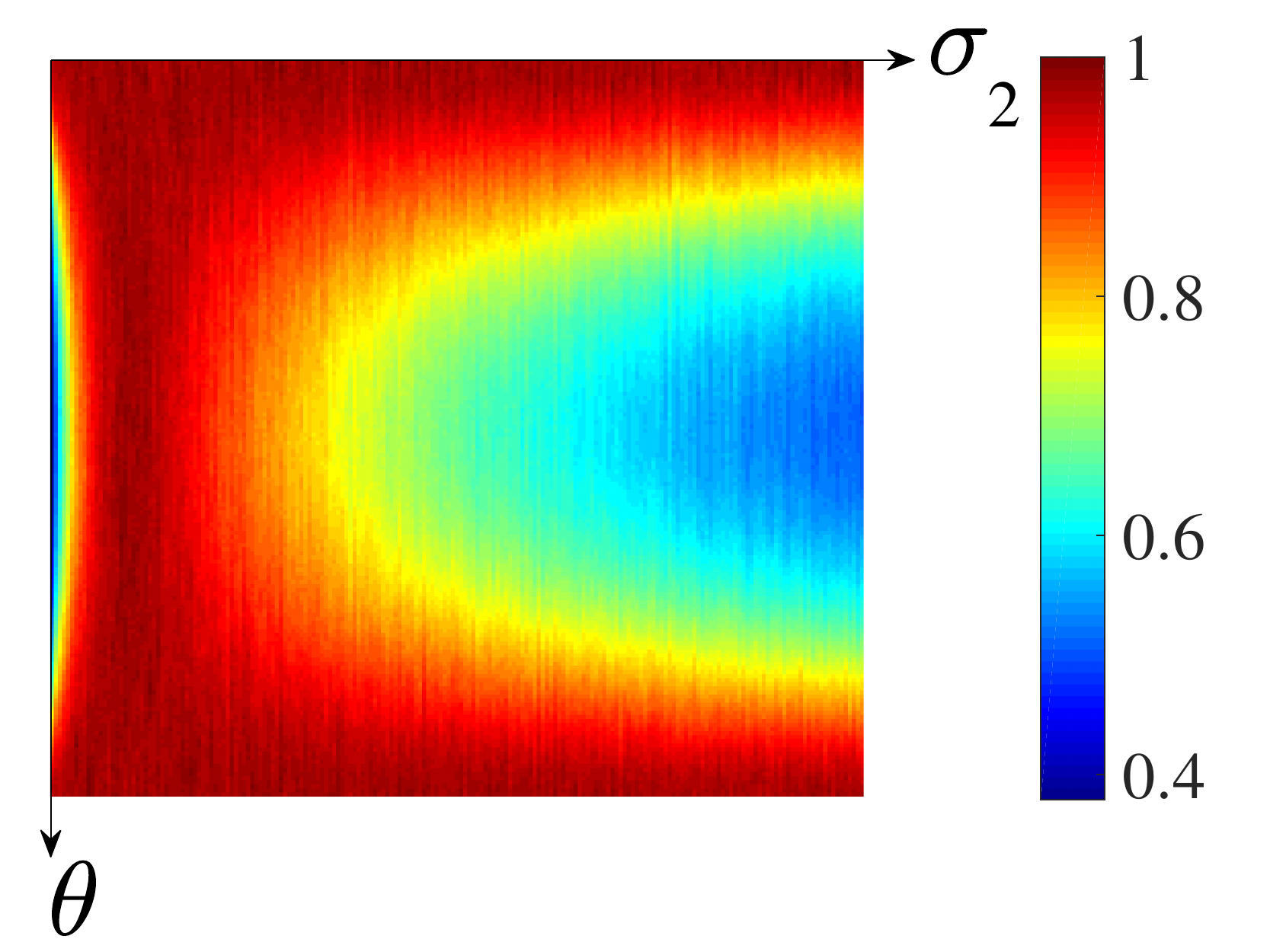}}
  \subfigure[$E(SDS)$]{
  \label{fig:rotation:a} 
  \includegraphics[width=0.22\linewidth]{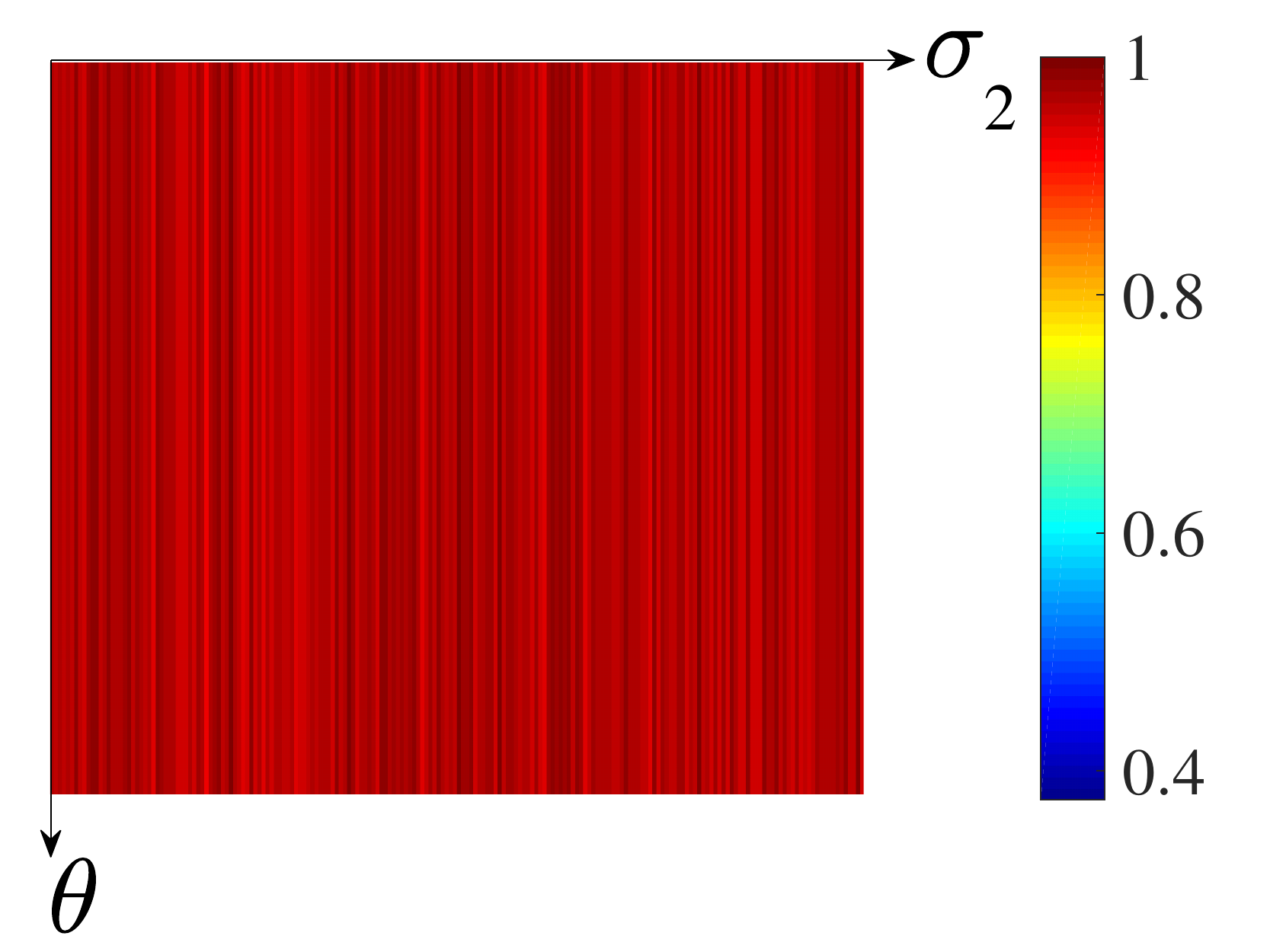}}
 \caption{The expectation maps of BBS and SDS in 2D Gaussian case with rotation. Points in $T$ and $Q$ are drawn from $N$, with $\mu=(0,0)$, $\sigma_1=1$, and $\sigma_2 \in (0, 10]$. $Q$ is further rotated by $\theta$, $\theta \in[0,-\pi]$. (a) shows an example of $T$ and (b) is generated by rotating sampled points. (c), (d) are the expectation maps of BBS and SDS respectively by varying $\theta$ and $\sigma_2$. It can be clearly observed that the expectation of SDS is almost invariant to rotation while BBS drops most when $T$ and $Q$ overlap least (i.e., $\theta=-\pi/2$).}
  \label{fig:rotation}
\end{figure*}

\noindent\textbf{Analysis of rotation robustness.} To show the robustness against rotation, we analyze the expectation of similarity between two sets $T$ and $Q$ drawn from 2D Gaussian models, as shown in Figure. \ref{fig:rotation}, we fix the parameters except $\theta$ and $\sigma_2$ to validate the effect of rotation angle along with the shape of the Gaussian. In the case of BBS, as we can observe from Figure. \ref{fig:rotation} (c), when $\sigma_2$ is extremely small, the points drawn are likely to form a line, which is sensitive to rotation as lines overlap little after rotation. This is also the case when $\sigma_2\gg\sigma_1$, as it can be observed that the expectation decreases gradually with the increase of $\sigma_2$. Also, isotropic Gaussian is supposed to be unaffected by the rotation, which can be convinced from Figure. \ref{fig:rotation} (c) that when $\sigma_1=\sigma_2=1$, the expectation keeps well with respect to the rotation. On the other hand, SDS shows the invariance to the rotation despite the shape change of distribution in Figure. \ref{fig:rotation} (d).



 \begin{figure*}[t]
 \centering
 \subfigure[Rotation]{
  \label{fig:AUC:a} 
  \includegraphics[width=0.23\linewidth]{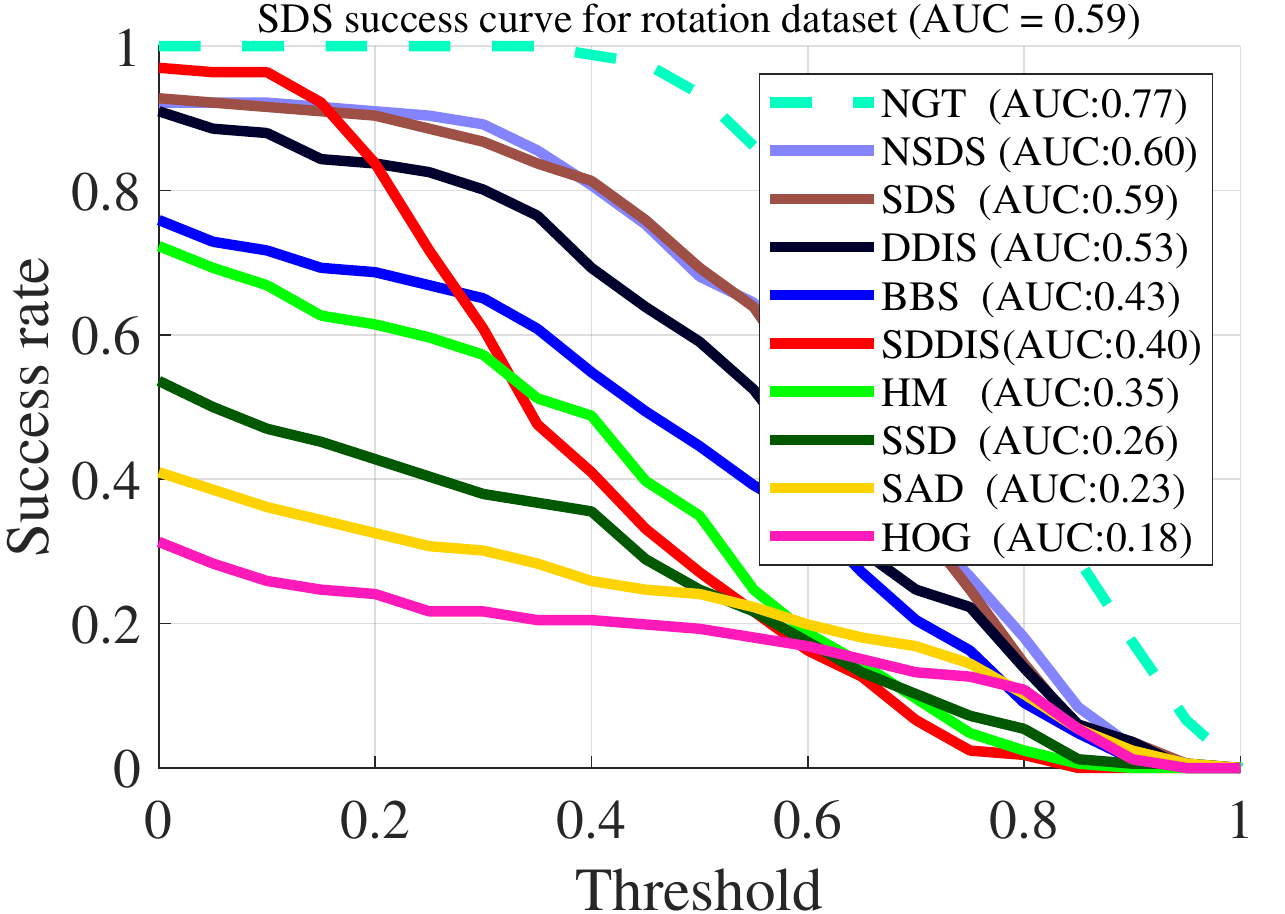}}
 \subfigure[Scaling]{
  \label{fig:AUC:b} 
  \includegraphics[width=0.23\linewidth]{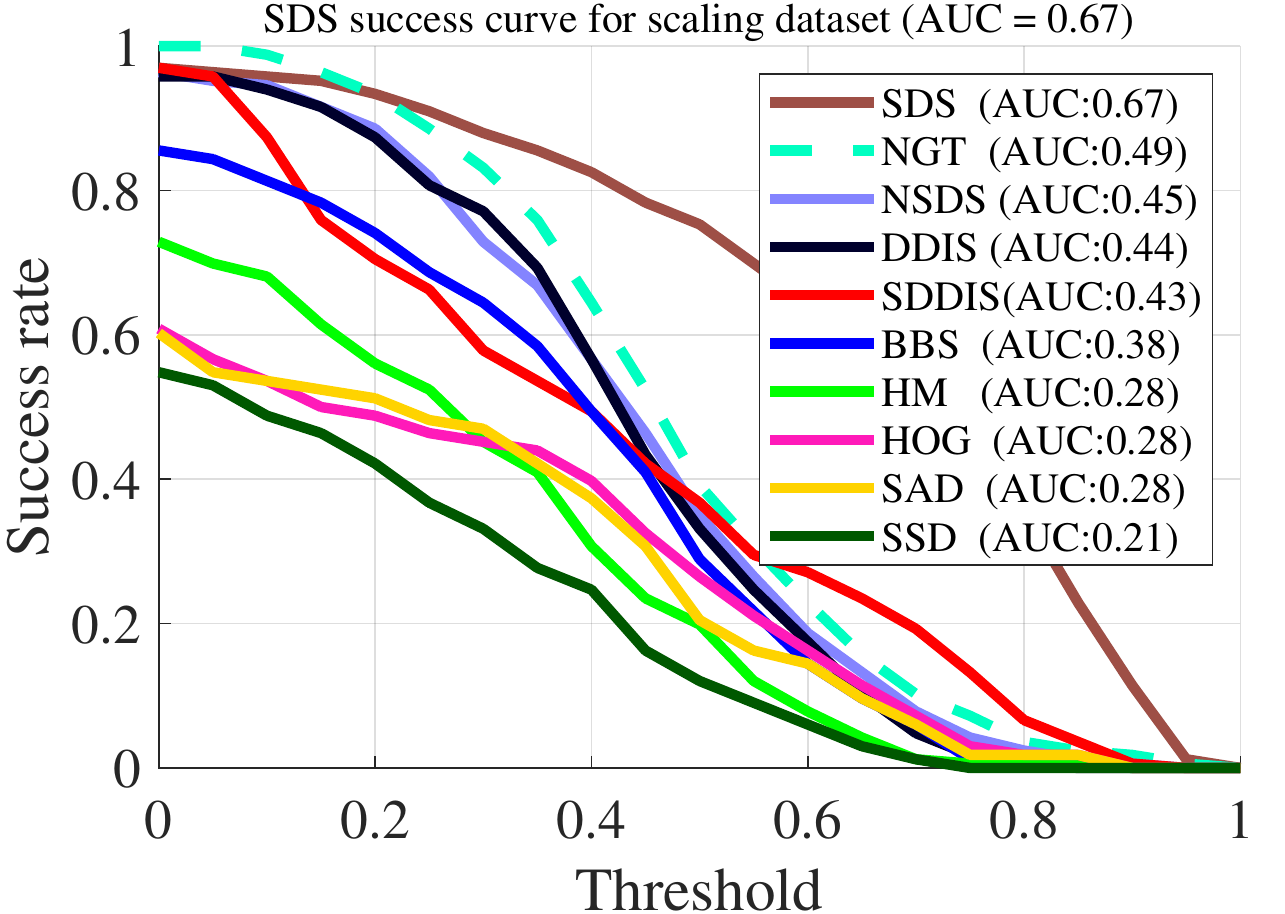}}
 \subfigure[Rotation-scaling]{
  \label{fig:AUC:c} 
  \includegraphics[width=0.23\linewidth]{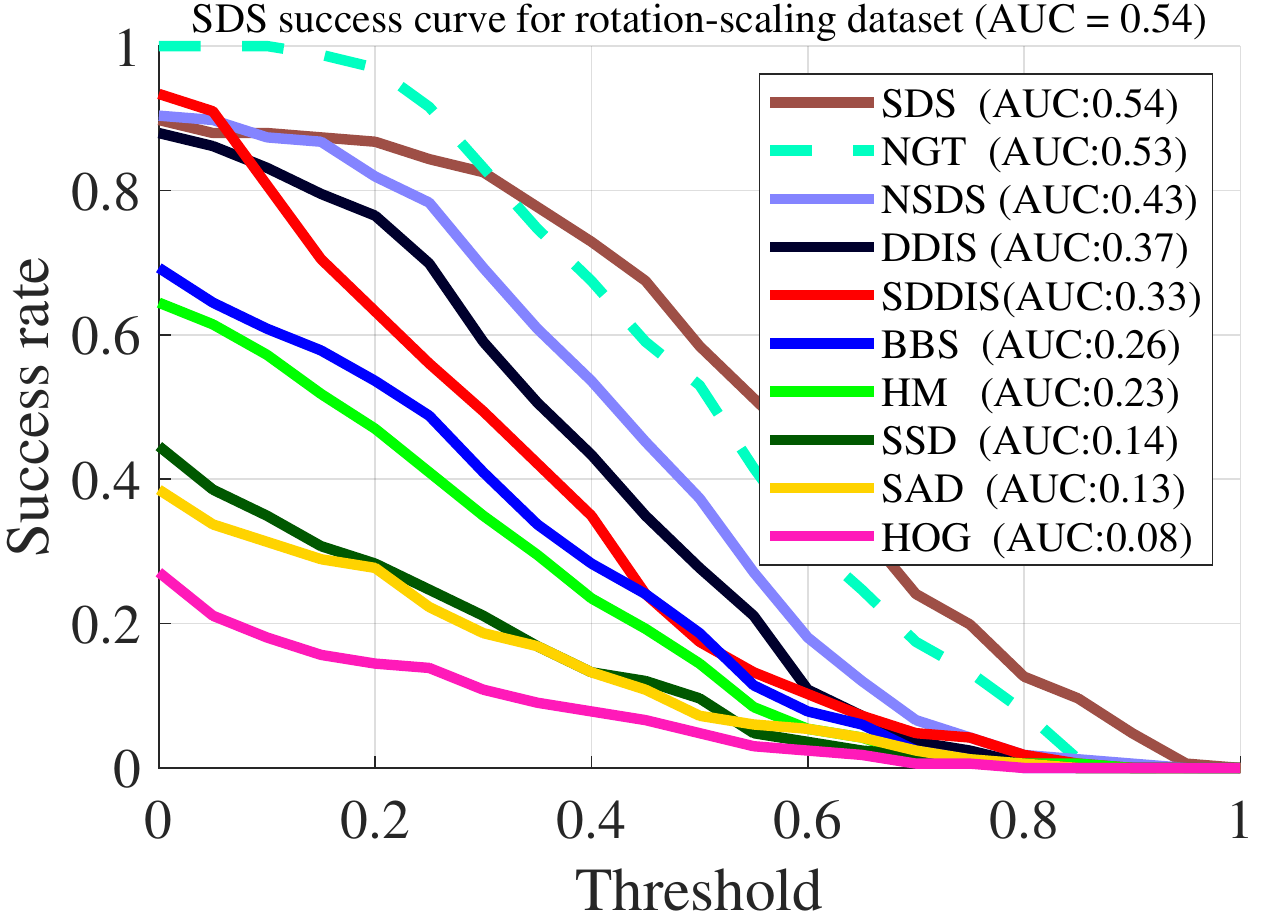}}
   \subfigure[All data]{
   \label{fig:AUC:d} 
 \includegraphics[width=0.23\linewidth]{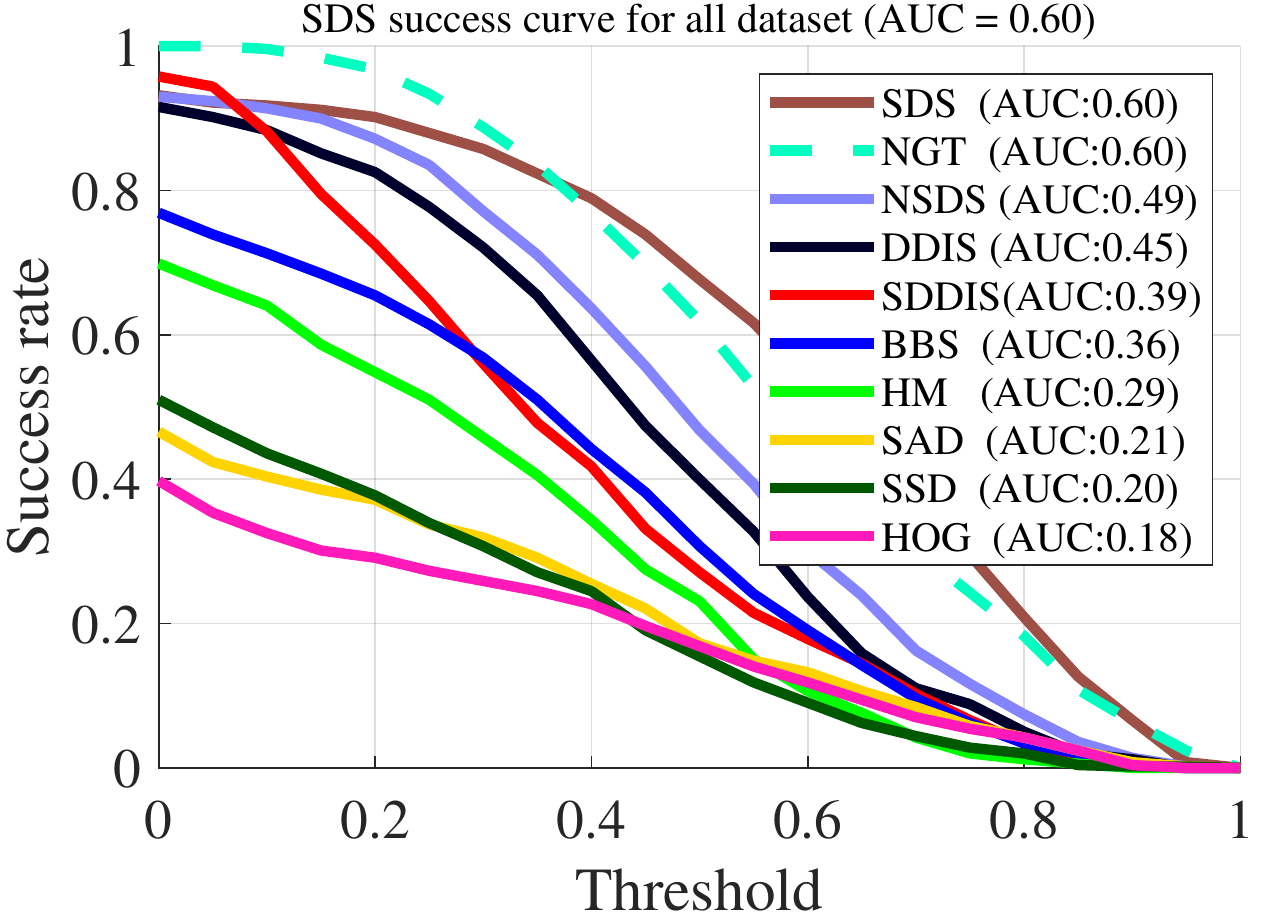}}
 \caption{Comparison on success rate with respect to the variation of the overlap rate threshold. SDS and SDDIS run in multi-scale and others are with fixed scales. The dotted curve (NGT) is the performance of the ground truth with fixed scales (i.e., results are represented by rectangles with centroids of the scale-variable ground truths + fixed sizes of the templates). AUC represents for the average success rate with respect to each curve. (a), (b) and (c) show the success curves over the rotation, scaling and rotation-scaling datasets, respectively. (d) shows the success curves by combining (a), (b), and (c). Best viewed in color.}
\label{fig:AUC}
\end{figure*}

\begin{figure*}[t]
 \centering
 \subfigure[Template]{
  \label{fig:example:a} 
  \includegraphics[width=0.17\linewidth]{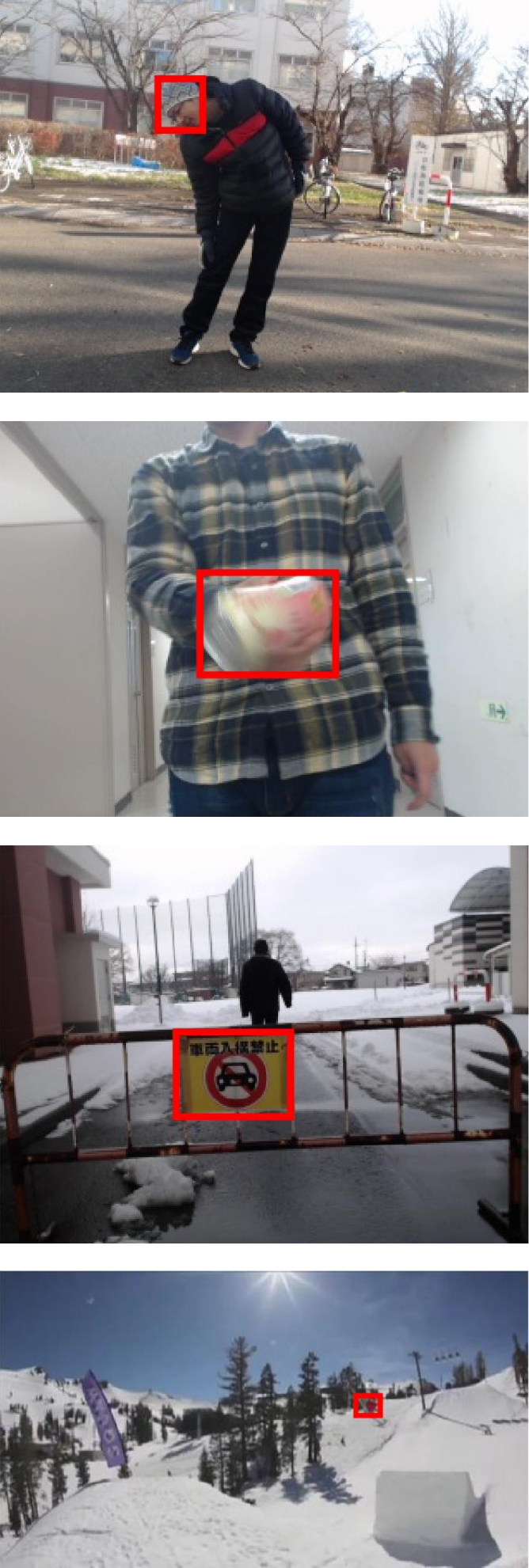}}
 \hspace{0.005in}
 \subfigure[Matching Results]{
  \label{fig:example:b} 
  \includegraphics[width=0.224\linewidth]{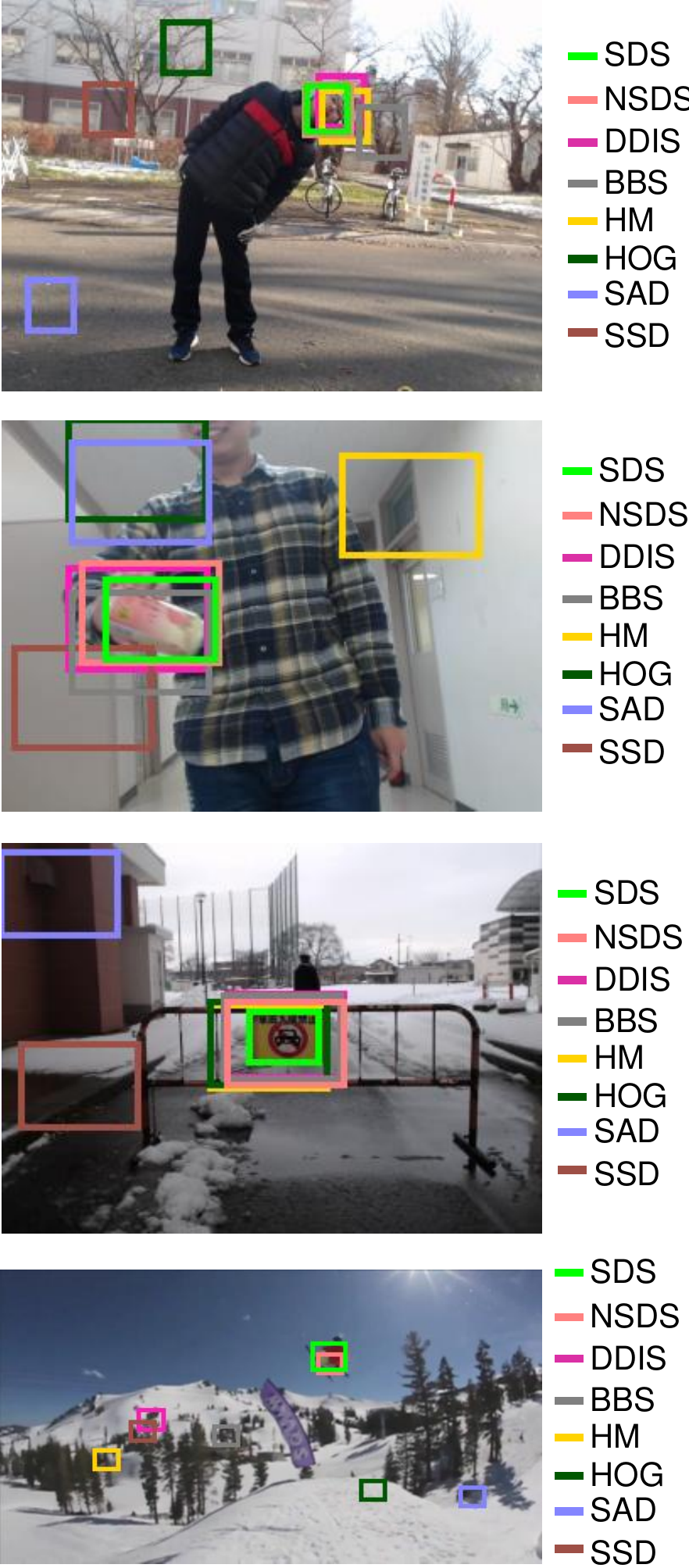}}
 \hspace{0.005in}
 \subfigure[SDS]{
  \label{fig:example:c} 
  \includegraphics[width=0.17\linewidth]{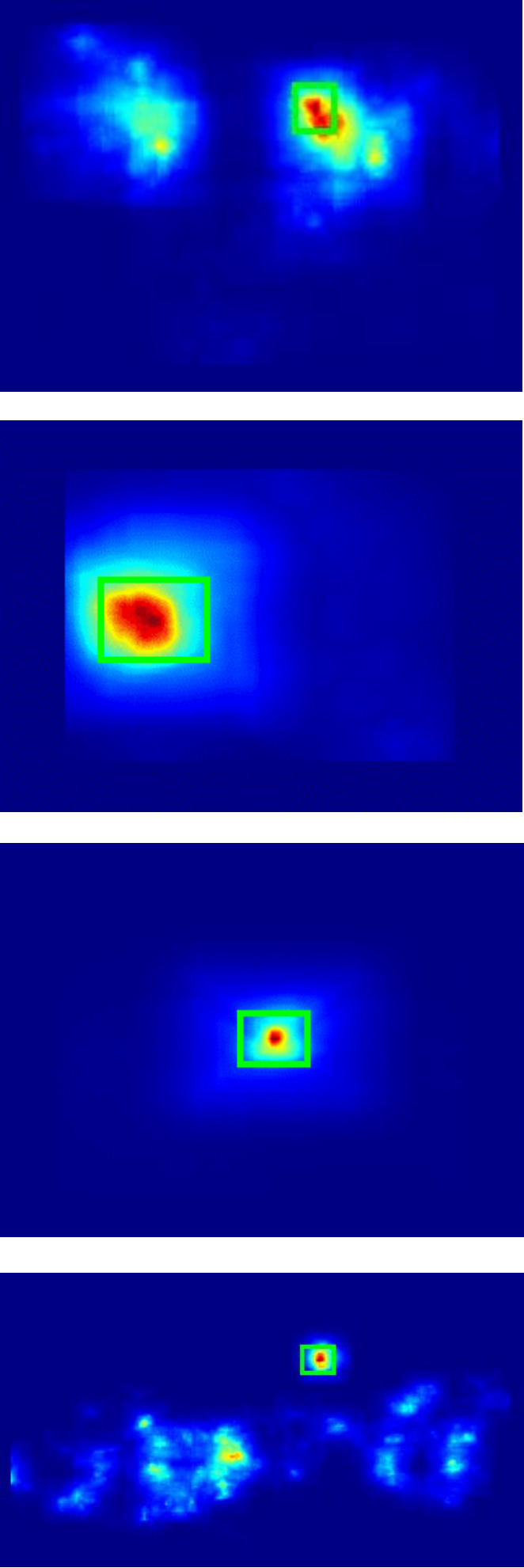}}
 \hspace{0.005in}
 \subfigure[NSDS]{
  \label{fig:example:c} 
  \includegraphics[width=0.17\linewidth]{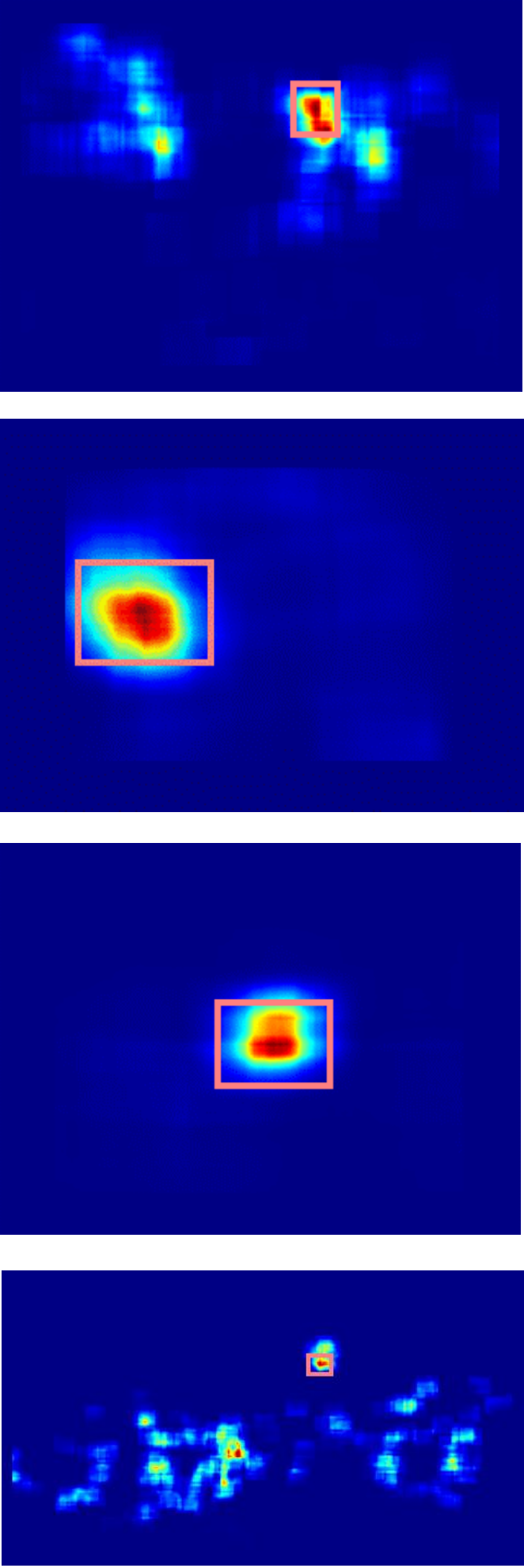}}
 \hspace{0.005in}
 \subfigure[DDIS]{
  \label{fig:example:d} 
  \includegraphics[width=0.17\linewidth]{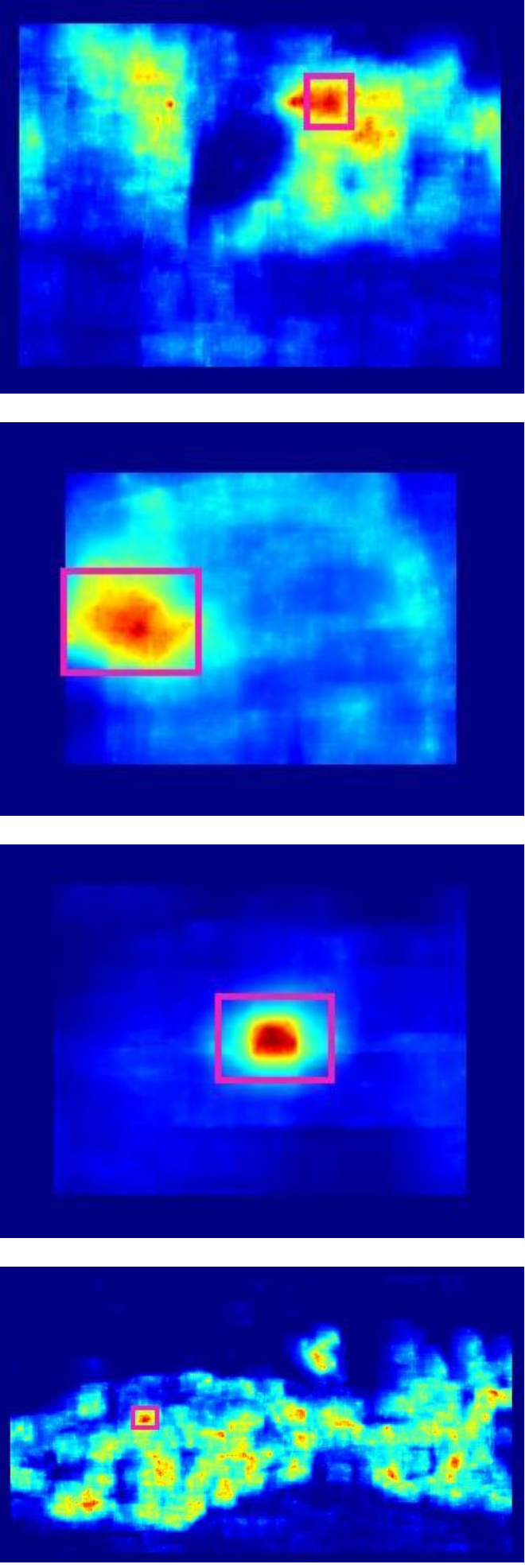}}
 \label{fig:example:e} 
 \caption{Examples of matching results. (a) Template is represented by a red rectangle. (b) Illustration of comparative results. (c)$\sim$(e) The likelihood maps of SDS, NSDS and DDIS, respectively. The candidate window with global maximum in each map is selected as the final matching result. In the likelihood map of SDS, every pixel has multiple similarity values due to multi-scale candidates, and only the maximum one is shown.}
 \label{fig:example} 
\end{figure*}

\section{Experiment Results}
We conduct a comprehensive 
experiment with both qualitative and quantitative tests to validate the superiority of SDS comparing with the state-of-the-art methods BBS \cite{dekel2015best,oron2018best} and DDIS \cite{talmi2017template}, as well as several conventional methods. We follow the same procedure as suggested in \cite{dekel2015best,talmi2017template} for a fair comparison. Note that as SDS can be employed with multi-scale windows, we simultaneously compare the performance of SDS with fixed scale, which is referred to as NSDS. In addition, similar to SDS, we also employed DDIS to the multi-scale candidate windows for comparison, denoted as SDDIS. 

Multiple datasets are utilized for comparison. We originally collected 42 videos 
under different unconstrained environments and extract frames to create a benchmark for evaluating the performance of template matching involving overall rotation and scaling on the object. Ground truths are scale-variable and annotated manually image by image. Besides, this benchmark also includes other challenges like complex deformations, occlusion, background clutter, etc. The benchmark is subdivided into three datasets: (1) rotation dataset, (2) scaling dataset and (3) rotation-scaling dataset for detail evaluation, each of them includes 166 reference-target image pairs, respectively. It is noteworthy that each dataset also includes other photometric and geometric transformations as they are taken under unconstrained environments. As to the evaluation criteria, following previous works\cite{dekel2015best,talmi2017template}, we employ the success ratio based on the overlap rate between ground truth $W_g$ and matching result $W_r$ to measure the accuracy, which is defined as: $\left | W_r\cap W_g \right |/\left |W_r\cup W_g \right |$. Here, the operator $\left | \cdot \right |$ is to count the number of pixels within a window.


We compare our proposed methods (SDS and NSDS) to DDIS and its multi-scale implementation SDDIS, BBS, HM, HOG, SAD, and SSD. The scaling factor with respect to both $x$ and $y$ axes range from 0.5 to 2, with step 0.1. The patch size of SDS, DDIS and BBS patch is fixed to $2\times 2$. We report the result in Figure. \ref{fig:AUC}. SDS/NSDS outperforms the other comparative methods with respect to the area-under-curve (AUC) score. NGT curve is to show the limitation of performance when calculating the success rate with fixed scales.
Matching examples are shown in Figure. \ref{fig:example}. 1st and 2nd rows show that SDS is robust against overall rotation. 3rd and 4th rows demonstrate that SDS can deal with scaling problem well. The likelihood maps show that SDS/NSDS is more distinct and yields in better-localized modes compared to other methods. 

\section{Conclusion}
We proposed a novel multi-scale template matching method in unconstrained environments, which is robust against overall scaling, intense rotation while taking advantage of global statistic based similarity measure to deal with complex deformations, occlusions, etc. Extended bidirectional diversity combined with rank based nearest neighbor search forms a scale-robust similarity measure, and the exploit of polar coordinate further improves the robustness against rotation. The experimental results have shown that SDS can remarkably outperform other competitive methods. On the other hand, SDS may fail when the template is too small to achieve a statistical score. The remained future work is to add a rotation parameter to the candidate windows to achieve rotation-specific matching results.

\bibliography{egbib}

\end{document}